\documentclass{article} % For LaTeX2e
\usepackage{iclr2027_conference,times}

\usepackage{amsmath,amsfonts,bm}

\def\eqref#1{equation~\ref{#1}}
\def\1{\bm{1}}

\DeclareMathAlphabet{\mathsfit}{\encodingdefault}{\sfdefault}{m}{sl}
\SetMathAlphabet{\mathsfit}{bold}{\encodingdefault}{\sfdefault}{bx}{n}

\usepackage{hyperref}
\usepackage{url}

\usepackage{graphicx} % includegraphics
\usepackage{wrapfig} % wrapigure

\usepackage{algorithmic}
\usepackage{algorithm}

\usepackage{booktabs}    % professional-quality tables

\usepackage{adjustbox} % for resizing tables

\usepackage{multirow}

\usepackage{amssymb}

\usepackage{subcaption}   % for the subfigure environment

\usepackage{tcolorbox}

\definecolor{inferencebg}{RGB}{230,240,255}   % light blue
\definecolor{trainingbg}{RGB}{255,230,230} % light red

\newtcolorbox{trainingexaggeration}{
  colback=trainingbg, colframe=trainingbg,
  boxrule=0pt, arc=2pt, left=2pt, right=2pt, top=2pt, bottom=2pt
} 
\newtcolorbox{inferenceexaggeration}{
  colback=inferencebg, colframe=inferencebg,
  boxrule=0pt, arc=2pt, left=2pt, right=2pt, top=2pt, bottom=2pt
}
\usepackage{caption}       % for \captionof in the Teaser image
\usepackage{tikz}          % pipeline schematic (Fig.~\ref{fig:pipeline})
\usetikzlibrary{arrows.meta,positioning}

\usepackage{amsthm}

\theoremstyle{definition}

\theoremstyle{remark}

\newcommand{\RNum}[1]{\uppercase\expandafter{\romannumeral #1\relax}}

\newcommand{\datasetname}{ExMoCap}

\title{Controllable Exaggeration for Generative Motion Models via Training-Time Adaptation and Inference-Time Guidance}

\author{
\parbox{\textwidth}{
Amirhossein Zamani$^{1,2,3}\;$ \thanks{Correspondence to: {\tt\small amirhossein.zamani@mila.quebec}} 
\hspace{5pt}
\qquad
Arianna Rampini $^{1}$\hspace{5pt}
\qquad
Bruno Roy $^{1}$\hspace{5pt}
}
\vspace{5pt}\\
$^1$ Autodesk Research \hspace{15pt} $^2$ Mila -- Quebec AI Institute \hspace{15pt} $^3$ Concordia University \vspace{10pt} \\
\hspace{1mm} \small Project page: \url{https://ahhhz975.github.io/ControllableMotionExaggeration/}
}

\iclrfinalcopy % COMMENTED FOR DOUBLE-BLIND SUBMISSION. Uncomment ONLY for the camera-ready version.

\begin{document}

\maketitle
%%%%%%%%%%%%%%%%%%%%%%%%%%%% Teaser Image %%%%%%%%%%%%%%%%%%%%%%%%%
\begin{figure}[h]
\vspace{-20pt} % tweak vertical spacing if needed
    \begin{center}
    \centerline{\includegraphics[width=1\textwidth]{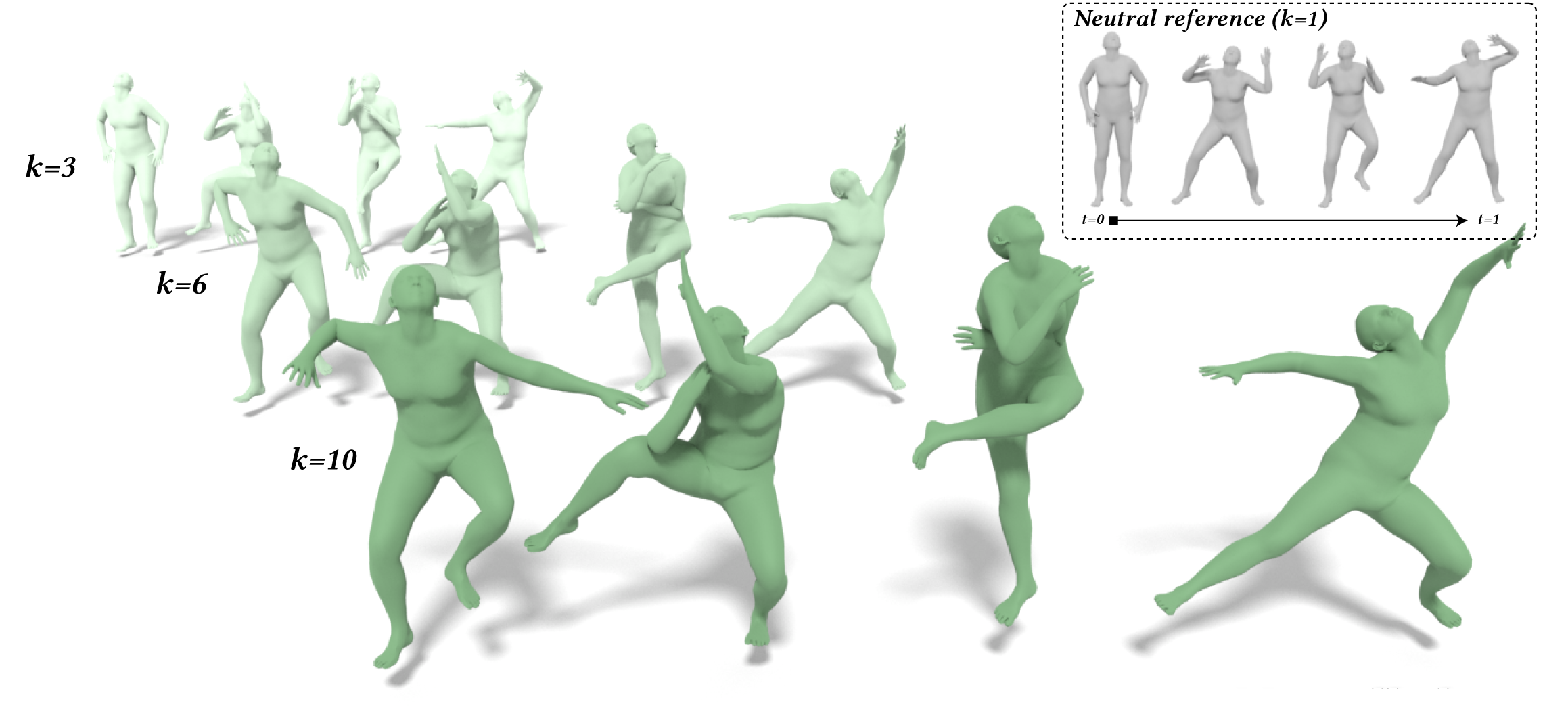}}
    \vspace{-10pt}
    \caption{Our approach generates controllable and plausible motion exaggerations ($\{k=3,k=6,k=10\}$ in green) from a neutral reference motion (dashed-line box in gray). Increasing $k$ progressively amplifies the dance motion while preserving its underlying action intent.}
    \label{fig:teaser}
    \end{center}
\end{figure}
%%%%%%%%%%%%%%%%%%%%%%%%%%%%%%%%%%%%%%%%%%%%%%%%%%%%%%%%%%%%%%%%

\vspace{-10pt}
\begin{abstract}
\vspace{-5pt}

    Recent motion generative models have demonstrated strong capabilities in synthesizing physically plausible character motion, but often overlook established animation principles used by professional animators to ground and design their animation work. Understanding and incorporating these principles into motion generative pipelines is essential for producing motions that serve not only physically grounded applications but also the needs of the character animation community. This enables the creation of characters that not only move in physically plausible ways but also feel alive, expressive, and engaging. To close this gap, we focus on the \textit{Exaggeration} principle of animation and investigate how it can be incorporated into modern motion generative pipelines to produce more expressive character motions. To this end, we introduce a framework that operates at two stages of existing motion generative pipelines. The first stage introduces exaggeration during training, where we perform supervised fine-tuning of pre-trained text-to-motion models on our curated exaggeration dataset. The second stage operates at inference time, where we: (i) introduce a mathematical formulation of exaggeration based on dynamic movement primitives (DMPs); and (ii) leverage this formulation as an exaggeration guidance signal to guide existing diffusion and flow-matching text-to-motion generation models toward exaggerated motion without additional training. Through qualitative and quantitative evaluations against three strong motion generation models, we show that our methods generate more exaggerated and expressive motions while preserving neutral reference motion intent and physical plausibility.

\end{abstract}

\vspace{-10pt}

\section{Introduction}
\label{Introduction}
{
    \vspace{-7pt}
     Character animation is a long-standing field in which artists and animators design and combine different techniques to create character motion that conveys dynamic behavior and interactions with the environment, as seen in cartoons, films, and games. As a common practice, animators rely on a set of well-established principles, commonly known as the 12 Principles of Animation, which were proposed by Disney animators in their book \textit{The Illusion of Life} \citep{AnimationPrinciples}. These principles provide a framework that guides animators in designing and refining character motion. Understanding and incorporating these principles into modern motion generation pipelines is essential for producing motions that are relevant not only to physically grounded applications but also to the character animation community \citep{3DComputerAnimation}. This integration could unlock more compelling generative character animation workflows, combining physically plausible motion with the expressiveness needed to make characters feel alive on screen. While recent motion generative foundation models \citep{MDM, Kimodo, HY-Motion, CloSD, AnyTop} produce realistic character motions, they often neglect the animation principles in the generation pipeline. Specifically, their focus is mainly on generating physically plausible motions that mimic real-world motion and match how humans and animals actually move. To close this gap, we target the \textit{Exaggeration} principle of animation, which is one of the most relevant principles for presenting motion in an expressive way \citep{AnimationBench, 3DComputerAnimation, AnimationPrinciples}, and aim to generate motions that are exaggerated and expressive. We provide a description of the 12 animation principles in Appendix~\ref{app:Remarks_12_AnimationPrinciples}. 

      % \vspace{-3pt}
     
     Incorporating the \textit{Exaggeration} principle into a motion generative pipeline requires understanding how animators interpret exaggeration and how they apply it in practice when designing expressive motions for characters. To understand such requirements, we had discussions with professional animators, which revealed several important insights that we account for in our pipeline design. As a result of these discussions, we identify three properties that an exaggerated motion generation method should satisfy: (i) \textit{controllable exaggeration}, where the animator should be able to adjust the motion exaggeration intensity through parameters that produce progressively more exaggerated motion; (ii) \textit{motion-intent preservation}, where exaggeration should amplify the expressive shape of the motion without changing the underlying action. For example, an exaggerated punch should still read as the same neutral punch action, only with stronger movement quality; and (iii) \textit{physical plausibility}, where increasing exaggeration should not introduce severe motion artifacts, such as invalid body poses, foot penetration, or unrealistic global movement. To address these requirements, we introduce a two-stage pipeline that incorporates exaggeration at the training and inference stages of existing motion generative foundation models. In the first stage, our \textit{training-time exaggeration} approach teaches pre-trained motion generative models to directly learn the characteristics of exaggerated motion from real motion capture performances with different exaggeration intensities through supervised fine-tuning of the pre-trained models. We study this approach across different motion generative architectures, including diffusion models \citep{Kimodo, MDM} and flow-matching models \citep{HY-Motion}, using an exaggeration dataset that we design and construct specifically for this task. In the second stage, our \textit{inference-time exaggeration} approach introduces exaggeration without modifying the parameters of the pre-trained model. Specifically, we mathematically formulate the \textit{Exaggeration} principle of animation using dynamic movement primitives (DMPs) \citep{DMP} and use this formulation to steer the generation process of text-to-motion models toward progressively exaggerated motions. 

    % \vspace{-3pt}
    
    The main contributions of this study are: \emph{(i)} To the best of our knowledge, this is the first work to formulate and incorporate an animation principle into modern motion generative foundation models (Sec.~\ref{sec:methodology}); \emph{(ii)} We present two complementary approaches for synthesizing exaggerated and expressive character motions, operating at training time and inference time of motion generative pipelines, respectively (Sec.~\ref{sec:training_time_exaggeration} and Sec.~\ref{sec:inference_time_exaggeration}); and \emph{(iii)} We develop an evaluation suite that includes both classifier-based and classifier-free approaches for evaluating exaggerated motions (Sec.~\ref{sec:quantitative_evaluation}).
}

\vspace{-5pt}

\section{Related Work}
\label{RelatedWork}
{
    \vspace{-5pt}
    
    \noindent\textbf{Expressive Motion Generation.} Existing approaches to expressive motion generation broadly focus on either characterizing expressiveness or manipulating motion style. A line of work seeks to quantify expressiveness so that it can be measured and controlled. Laban Movement Analysis \citep{Laban} is the foundation of this line of work: it maps raw kinematics to interpretable descriptors of effort and shape \citep{MotionDescriptors, Laban, LabanEmotionAnalysis, RecognizeEmotionFromDance}, providing per-joint and per-frame measures of quantity of motion, weight, time, and spatial extent. Another line of work focuses on motion style editing, including unpaired motion style transfer \citep{Aberman2020}, body-part-level style control \citep{MotionPuzzle2022}, and, most closely related to our work, procedural squash-and-stretch and smear effects \citep{Smear2024}. These methods either transfer a reference style or apply hand-authored effects, but they do not specifically model any animation principles such as exaggeration that we target in this work.

    % \vspace{-5pt}

    \noindent\textbf{Motion Generative Models.} Research on motion generative models has grown rapidly in recent years \cite{khani2025motion, zhu2023human}. Text- and kinematics-conditioned generative models produce whole-body motion from natural-language prompts \citep{Kimodo, HY-Motion, MDM, AnyTop}. They offer control through text and an extensive set of kinematic constraints, including full-body pose keyframes, end-effector targets, and 2D paths, while operating on carefully designed motion representations. While the motions synthesized by these generative models are mostly physically plausible and realistic, they may not be suitable for character animation applications, where characters should feel alive and engaging. In this work, we leverage the neutral motions synthesized by these motion foundation models as reference motions and aim to make them more expressive and exaggerated.

    % \vspace{-5pt}

    \noindent\textbf{Dynamic Movement Primitives.} DMPs \citep{DMP, DMP_Survey} model a movement as a stable second-order dynamical system. DMPs have been widely used in robotics for imitation learning \citep{DMP, DMP_ImitationLearning}, generalization \citep{DMP_Generalization}, and, most relevant to our work, expressive robot motion \citep{DMP_Robotics}. In the context of character and robot motion, where rotations live on a non-Euclidean manifold, DMPs have also been formulated for quaternions \citep{Ude2014, Koutras2020}. In our work, we fit DMPs either on Euler channels, on the continuous 6D rotation representation \citep{Zhou2019}, or directly on the motion generator's native representation. To our knowledge, modeling the \textit{Exaggeration} principle of animation using DMPs has been explored only for single start-goal robot trajectories in \citep{DMP_Robotics}, but not for multi-keyframe 3D character animation. In this work, we leverage DMPs and motion generative models to close this gap.
}

\vspace{-5pt}

\section{Methodology}
\label{sec:methodology}
{
    \vspace{-5pt}

    \begin{figure*}      
        \vspace{-17pt}
        \begin{center}
        \centerline{\includegraphics[width=1\linewidth]{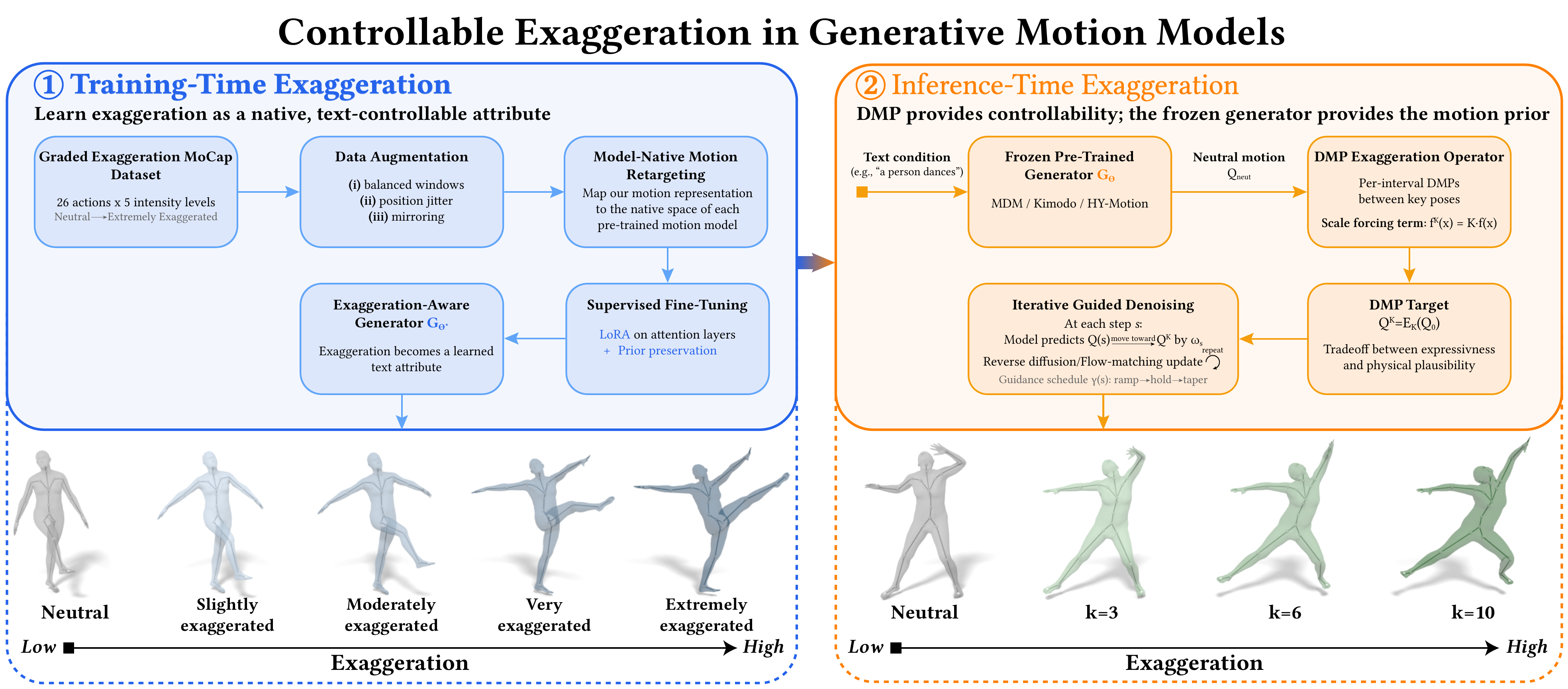}}
        % \vspace{-5pt}
        \caption{\small \textbf{An overview of the controllable motion exaggeration framework,} consisting of two stages: \emph{(1) \textcolor{blue}{Training-Time Exaggeration}}, which constructs a graded motion-capture dataset with 26 actions and five exaggeration levels, augment and retarget the data, and perform supervised fine-tuning with LoRA and prior preservation to learn exaggeration as a controllable, text-conditioned attribute (Sec.~\ref{sec:training_time_exaggeration}); \emph{(2) \textcolor{orange}{Inference-Time Exaggeration}}, where a frozen pre-trained motion generation model first produces a neutral motion $Q_{\mathrm{neut}}$, which is transformed by a DMP-based exaggeration operator $\mathcal{E}_K$ into an exaggerated target $Q_K$. Then, iterative guidance steers the sampling process toward $Q_K$, while the frozen generator regularizes the motion through its learned physical prior (Sec.~\ref{sec:inference_time_exaggeration}). The bottom examples show progressively increasing levels of exaggeration.}
        \label{methodology_overview}
        \end{center}
        \vspace{-15pt}
    \end{figure*}
        
    We introduce a framework for incorporating the \textit{Exaggeration} principle of animation into generative motion models at two stages of the generative pipeline. First, we introduce exaggeration during training through supervised fine-tuning of pre-trained text-to-motion models on an exaggeration dataset that we construct, containing 26 actions, with five exaggeration levels per action, enabling the models to learn exaggeration as a controllable text property. Second, at inference time, we steer pre-trained motion generation models toward exaggerated motions without modifying their parameters. This stage consists of two steps: \emph{(i)} formulating exaggeration mathematically using dynamic movement primitives (DMPs) \citep{DMP}, and \emph{(ii)} using this formulation as a guidance signal for diffusion \citep{Kimodo, MDM} and flow-matching \citep{HY-Motion} text-to-motion models. An overview of the training-time and inference-time procedures is shown in Fig.~\ref{methodology_overview} and Algorithm~\ref{alg:exaggeration}.

    % \vspace{-5pt}

    \subsection{Problem Statement and Preliminaries}    
        % \vspace{-3pt}
        \label{preliminary}
        
        Let a character motion be represented as a sequence $Q={q_1,\dots,q_T}$, where $q_t$ denotes the character skeleton representation at frame $t$. 

        \begin{wrapfigure}[23]{r}{0.5\linewidth}
            \vspace{-28pt}
            \begin{minipage}{1\linewidth}
            \begin{algorithm}[H]
              \caption{Controllable Motion Exaggeration}
              \label{alg:exaggeration}
              \scriptsize
              \begin{algorithmic}[1]
            
                \begin{trainingexaggeration}
                \STATE \textbf{Stage 1: Training-Time Exaggeration (Sec.~\ref{sec:training_time_exaggeration})}
                \STATE Retarget the \datasetname ~ dataset
                $\mathcal{D}_{\mathrm{exag}}
                =\{(Q^{(\ell)},c_\ell)\}_{\ell=0}^{4}$
                \STATE Apply motion augmentation 
                \STATE Initialize $\theta\leftarrow\theta_0$ and insert LoRA
                adapters of rank $r$
                \FOR{$n=1$ \textbf{to} $N$}
                  \STATE Sample $(Q^{(\ell)},c_\ell)$, diffusion step $s$,
                  and noise $\epsilon$
                  \STATE Compute the diffusion training loss
                  $\mathcal{L}_{\mathrm{SFT}}$ using
                  $\mathcal{G}_{\theta}$
                  \STATE Compute the prior-preservation loss
                  $\mathcal{L}_{\mathrm{prior}}$
                  \STATE $\mathcal{L}\leftarrow
                  \mathcal{L}_{\mathrm{SFT}}
                  +\lambda\,\mathcal{L}_{\mathrm{prior}}$
                  \STATE Update the trainable LoRA parameters:
                  $\theta\leftarrow\theta-\eta\nabla_\theta\mathcal{L}$
                \ENDFOR
                \STATE $\theta^\star\leftarrow\theta$
                \hfill $\triangleright$ fine-tuned exaggeration model
                \end{trainingexaggeration}
            
                \begin{inferenceexaggeration}
                \STATE \textbf{Stage 2: Inference-Time Exaggeration (Sec.~\ref{sec:inference_time_exaggeration})}
                \STATE Set $\theta_{\mathrm{inf}}\leftarrow\theta_0$
                for TF, or $\theta_{\mathrm{inf}}\leftarrow\theta^\star$
                for SFT+TF
                \STATE $z_S\sim\mathcal{N}(0,I)$; sample neutral motion
                $Q_{\mathrm{neut}}$
                \STATE $Q_K\leftarrow\mathcal{E}_K(Q_{\mathrm{neut}})$
                \hfill $\triangleright$ DMP-exaggerated target
                \STATE Reset $z_S$ using the same seed as $Q_{\mathrm{neut}}$
                \FOR{$s=S$ \textbf{to} $1$}
                  \STATE $\widehat{Q}^{(s)}
                  \leftarrow\mathcal{G}_{\theta_{\mathrm{inf}}}
                  (z_s,s,c)$
                  \hfill $\triangleright$ clean-motion prediction
                  \STATE $\omega_s\leftarrow\omega\,\gamma(s)$
                  \STATE $Q_e^{(s)}
                  \leftarrow\widehat{Q}^{(s)}
                  +\omega_s\!\left(Q_K-\widehat{Q}^{(s)}\right)$                  
                  \STATE $Q_e^{(s),\mathrm{root}}
                  \leftarrow Q_0^{\mathrm{root}}$
                  \hfill $\triangleright$ root lock
                  \STATE $z_{s-1}\leftarrow
                  \mathrm{DDIM}(z_s,Q_e^{(s)})$
                \ENDFOR
                \STATE \textbf{return} $Q_e^{(0)}$
            
                \end{inferenceexaggeration}
            
              \end{algorithmic}              
            \end{algorithm}            
            \end{minipage}            
        \end{wrapfigure}        

        \vspace{-4pt}
        
        We consider a conditional motion generative model $\mathcal{G}_{\theta}$ with parameters $\theta$ and an optional semantic condition $c$, such as a text prompt. A neutral motion generated by the pre-trained model is $Q_{\mathrm{neut}}=\mathcal{G}_{\theta}(c)$. Given an exaggeration intensity $K$, our goal is to generate $Q_K$ whose expressive characteristics increase with $K$ while preserving the semantic content and physical properties of $Q_{\mathrm{neut}}$. Specifically, exaggeration should provide: (i) \textit{controllable exaggeration}, where $K=1$ recovers the neutral motion and $K>1$ progressively increases exaggeration; (ii) \textit{motion-intent preservation}, where exaggeration changes \textit{how} an action is performed without changing \textit{what} action is performed; and (iii) \textit{physical plausibility}, where increasing exaggeration does not introduce severe artifacts such as invalid body poses, foot penetration, or unrealistic global movement.

    % \vspace{-5pt}

    \subsection{Training-Time Exaggeration}
    \label{sec:training_time_exaggeration}
    {
        \vspace{-3pt}

    % dataset
    \textbf{Dataset Construction.} One way to model exaggeration is to learn how humans actually perform exaggerated motions. To this end, we recruited two professional actors to perform different exaggerated and expressive movements and constructed a motion-capture dataset containing $26$ actions, with each action performed at five different levels of exaggeration. We refer to this dataset as \textbf{\datasetname} \footnote{ExMoCap was created by Autodesk Research specifically for this study under its established data-governance and ethical-research practices and is proprietary and not publicly available.}. We denote the exaggeration level by $\ell\in\{0,1,2,3,4\},$ where $\ell=0$ corresponds to the neutral version of the action and $\ell=4$ corresponds to its most exaggerated version. This graded structure provides supervision for learning how the movement quality of an action changes as exaggeration increases while its motion intent remains unchanged. 

    % \vspace{-3pt}
    
    % SFT formulation
    \textbf{SFT Formulation.} Then, we perform supervised fine-tuning of pre-trained motion generation models using \datasetname, where exaggeration intensity is introduced through the text conditioning. Specifically, each discrete level \(\ell\) is associated with an intensity description ranging from neutral to extremely exaggerated and appended to the underlying action caption (e.g., `\textit{a person dances, very exaggerated}''). This allows exaggeration to be learned as a text-conditioned attribute and enables users to control exaggeration at inference time by changing the intensity description while keeping the action unchanged. We fine-tune three pre-trained motion foundation models: MDM \citep{MDM}, HY-Motion \citep{HY-Motion}, and KIMODO \citep{Kimodo}. Since these models use different motion representations, we first retarget each motion sample in \datasetname~to a common SMPL-22 body skeleton represented by $3$D joint trajectories \citep{SMPL:2015}. Then, we resample and encode each trajectory separately in the native motion representation of each backbone, allowing every model to be fine-tuned in the same feature space and normalization used during its pre-training. Appendix~\ref{app:motion_representations_backbones} and Table~\ref{table:motion_representations_backbones} summarize these representations. Let \(Q^{(\ell)}\in\mathbb{R}^{T\times D}\) denote a motion at exaggeration level \(\ell\), with \(T\) frames and \(D=263\) for MDM’s HumanML3D motion representation, and let \(c_{\ell}\) denote its corresponding action-exaggeration text condition (e.g., `\textit{a person dances, very exaggerated}''). During fine-tuning of MDM \citep{MDM}, we sample a diffusion timestep \(s\) and Gaussian noise \(\epsilon\sim\mathcal{N}(0,I)\), and construct the corresponding noisy motion \(z_s\) as:

    \vspace{-10pt}
    \begin{equation}
        \label{eq:forward}
        z_s
        =
        \sqrt{\bar{\alpha}_s}\,Q^{(\ell)}
        +
        \sqrt{1-\bar{\alpha}_s}\,\epsilon,
        \qquad
        \bar{\alpha}_s=\prod_{u=1}^{s}(1-\beta_u),
    \end{equation}
    \vspace{-10pt}
        
        where $\{\beta_u\}$ is the diffusion noise schedule inherited from the pre-trained checkpoint. Then, in the reverse diffusion process, the model
        $\mathcal{G}_{\theta}(z_s,s,c_{\ell})$ predicts the clean motion $\hat{Q}^{(\ell)}$ from the noisy intermediate variable $z_s$: 
        $ \hat{Q}^{(\ell)} = \mathcal{G}_{\theta}(z_s,s,c_{\ell})$. We fine-tune the model using the clean-motion reconstruction objective of MDM~\citep{MDM}:
        
        \begin{equation}
            \label{eq:sft_diffusion_loss}
            \mathcal{L}_{\mathrm{SFT}}
            =
            \mathbb{E}_{Q^{(\ell)},s,\epsilon}
            \left[
            \frac{1}{D\sum_t M_t}
            \sum_{t=1}^{T}\sum_{d=1}^{D}
            M_t
            \left(
                Q^{(\ell)}-\mathcal{G}_{\theta}(z_s,s,c_{\ell})
            \right)_{t,d}^{2}
            \right],
        \end{equation}
        
        where $M$ is a validity mask excluding padded frames. The formulation for supervised fine-tuning of HY-Motion and KIMODO, which use flow-matching and diffusion-based generative formulations, respectively, is provided in Appendices~\ref{app:sft_formulation_hymotion} and~\ref{app:sft_formulation_kimodo}. We refer to this approach as \textit{supervised fine-tuning (SFT)} throughout the paper.

        % \vspace{-3pt}
        
        \textbf{LoRA and Prior-Preservation For Feasible Fine-Tuning.} Since our exaggeration dataset is considerably smaller than the datasets used to pre-train the motion foundation models, naively fine-tuning all model parameters $\theta$ can lead to overfitting and catastrophic forgetting \citep{CatastrophicForgetting, CatastrophicForgetting1}, causing the model to lose its ability to generate diverse motions during fine-tuning (Table~\ref{tab:mdm_sft_ablation}). To address this issue, inspired by \citep{LoRA, DreamBooth, LoRA-MDM}, we adopt three techniques. First, \textit{data augmentation}: we augment the captured motions using left-right mirroring, balanced windowing, and position jittering, significantly increasing the number of samples in \datasetname. Details of data augmentation techniques are provided in Appendix~\ref{app:sft_data_augmentation}. Second, \textit{low-rank adaptation (LoRA)}: we freeze the original model weights and learn only low-rank updates to selected linear layers of each backbone \citep{LoRA}, restricting the adaptation task to a smaller parameter space and helping retain the original motion prior. Third, \textit{prior preservation} \citep{DreamBooth}: fine-tuning batches additionally include motions from the original pre-training dataset, using the same denoising objective as Eq.~\ref{eq:sft_diffusion_loss}. 
        % This allows the model to learn exaggerated motions from \datasetname while preserving the motion prior learned from the original pre-training data. The resulting fine-tuned generator directly models exaggerated motion as $Q_{\ell}\sim p_{\theta^{\star}}(Q\mid c_\ell)$, where $\theta^{\star}$ denotes the fine-tuned parameters.

    }

    % \vspace{-3pt}
    
    \subsection{Inference-Time Exaggeration}
    \label{sec:inference_time_exaggeration}
    {
        \vspace{-3pt}
            
        Another way to generate expressive motions is to formulate and incorporate the \textit{Exaggeration} principle of animation into a pre-trained generative model through its inference process. The idea is to manipulate the diffusion or flow-matching process of motion generative models toward generating more exaggerated motions. However, this raises a natural question: what is the target exaggerated motion toward which the inference sampling process should be guided? To answer this question, in the next subsections, we propose an exaggeration operator $\mathcal{E}_{K}$ to formulate exaggeration and leverage this formulation as a guidance signal inside the generative sampling process during inference.

        % \vspace{-3pt}
        
        \subsubsection{Mathematical Formulation of Exaggeration}
        \label{sec:exaggeration_math_formulation}
        {
            \vspace{-3pt}            
            
            We propose to formulate the exaggeration as a geometric transformation of a neutral motion. Specifically, given a neutral motion $Q_{\mathrm{neut}}$, we define an exaggeration operator, $Q_K = \mathcal{E}_{K}(Q_{\mathrm{neut}})$, where $K$ is a scalar controlling the exaggeration. We design $\mathcal{E}_{K}$ such that $K=1$ reconstructs the neutral motion and $K>1$ progressively exaggerates its movement characteristics while preserving the underlying action. To this end, we leverage dynamic movement primitives (DMP) \cite{DMP}, presented below.
            
            % \vspace{-5pt}
            
            \textbf{Dynamic Movement Primitives Formulation.} DMPs represent a movement as a stable dynamical system that combines \emph{(i)} a goal-directed spring-damper term, which drives the motion from a start state to a goal state and thereby guarantees that the goal is reached, and \emph{(ii)} a learned forcing term, $f(x)$, which captures the shape, swings, and arcs the motion makes on its way to the goal. In this work, we use the DMP formulation of \citet{DMP}, in which, for a single motion channel $y$, the DMP transformation system is given by the following differential equation:

            \vspace{-8pt}
            
            \begin{equation}
                \label{eq:dmp_diff_equation}
                \tau^2 \ddot{y} = \alpha\big(\beta(g-y)-\tau\dot{y}\big)+f(x),
            \end{equation}

            % \vspace{-5pt}
             
            where $y_0$ and $g$ denote the start and goal states, $\alpha$ and $\beta$ are gains that control the spring-damper dynamics (critically damped for $\beta=\alpha/4$), $\tau$ controls the temporal scale, and $x$ is a phase variable that monotonically decays from $1$ to $0$,  $\tau\dot{x}=-\alpha_x x$, $x(0)=1$. The nonlinear forcing term $f(x)$ can be represented using a set of Gaussian basis functions $\psi_i(x)$,

            \vspace{-8pt}
            
            \begin{equation}
                \label{eq:dmp_forcing_term}
                f(x)
                =
                \frac{\sum_{i=1}^{N_w}\psi_i(x)w_i}{\sum_{i=1}^{N_w}\psi_i(x)}\;x(g-y_0),
            \end{equation}

            % \vspace{-5pt}
            
            where $\psi_i(x)=\exp(-h_i(x-c_i)^2)$ and $w_i$ denotes the learnable weight of basis function $i$, which is obtained in closed form using locally weighted regression \citep{LWR, LWR_Control} such that $f(x)\approx f_{\mathrm{target}}(t)$. This formulation provides an interpretation for exaggeration. Specifically, manipulating the forcing term allows us to amplify the expressive shape of a trajectory.
            % while leaving its bounding poses unchanged. 

            % \vspace{-5pt}
            
            \textbf{Per-Interval DMP Motion In-Betweening.} Conventional DMPs, which have been studied primarily in robotics \citep{DMP, DMP_Generalization, DMP_ImitationLearning, DMP_Robotics, DMP_Survey}, describe a single start-goal pair. However, a character motion violates this assumption because a motion sequence contains multiple key poses, i.e., multiple start-goal pairs, and naively fitting a single DMP to the entire motion sequence fails to accurately reconstruct the motion. To address this issue, we fit one DMP independently to every keyframe interval $[t_i,t_{i+1}]$ in a motion sequence. For each interval and each motion channel, we set $y_0$ and $g$ to the corresponding values of the two bounding poses. Given the neutral motion trajectory $y_{\mathrm{neut}}$ obtained from the pre-trained motion model, the target forcing term can be learned by rearranging Eq.~\ref{eq:dmp_diff_equation},

            \vspace{-8pt}
            
            \begin{equation}
                \label{eq:dmp_reorganized_diff_equation}
                f_{\mathrm{target}}(t)
                =
                \tau^2\ddot{y}_{\mathrm{neut}}(t)
                -
                \alpha\big(\beta(g-y_{\mathrm{neut}}(t))-\tau\dot{y}_{\mathrm{neut}}(t)\big).
            \end{equation}

            % \vspace{-5pt}
            
            Then, we estimate the basis weights $w_i$ in Eq.~\ref{eq:dmp_forcing_term} such that $f(x)$ reproduces $f_{\mathrm{target}}$. Since one DMP is fitted per keyframe interval, naively resetting each DMP with zero velocity introduces a ``stop-and-go'' effect at interval boundaries. We avoid this by initializing each interval with the terminal velocity of the preceding interval. This maintains velocity continuity through the key poses.
            
            % \vspace{-4pt}

            \textbf{Exaggerating Motion using DMP.} Learning the forcing term $f(x)$ in Eq.~\ref{eq:dmp_forcing_term}, extracts the shape of motion for every joint in the character skeleton. To exaggerate this motion, inspired by \cite{DMP_Robotics}, one way is to scale the forcing term by a factor,
            
            \vspace{-8pt}
            
            \begin{equation}
                \label{eq:dmp_scaled_forcing_term}
                f^{K}(x)=Kf(x).
            \end{equation}

            % \vspace{-5pt}
            
            Then, by replacing the forcing term in Eq.~\ref{eq:dmp_diff_equation} with the scaled forcing term from Eq.~\ref{eq:dmp_scaled_forcing_term}, we obtain:

            \vspace{-8pt}

            \begin{equation}
                \label{eq:dmp_scaled_diff_equation}
                \tau^2 \ddot{y}_{\mathrm{exag}} = \alpha\big(\beta(g-y_{\mathrm{exag}})-\tau\dot{y}_{\mathrm{exag}}\big)+Kf(x),
            \end{equation}

            % \vspace{-5pt}

            Then, given the scaled forcing term $Kf(x)$, we solve Eq.~\ref{eq:dmp_scaled_diff_equation} to obtain the updated exaggerated motion trajectory, including the joint positions $y_{\mathrm{exag}}$, velocities $\dot{y}_{\mathrm{exag}}$, and accelerations $\ddot{y}_{\mathrm{exag}}$. Then, the exaggerated motion is the updated joint positions $y_{\mathrm{exag}}$ obtained from this process. The entire process can be summarized as: 

            \vspace{-20pt}
            
            \begin{equation}
                \label{eq:dmp_exag_operator}
                Q_K=\mathcal{E}_{K}(Q_{\mathrm{neut}}), 
            \end{equation}    

            \vspace{-5pt}
            
            where $Q_{\mathrm{neut}}$ is the neutral motion, $\mathcal{E}_{K}$ represents all the steps from Eq.~\ref{eq:dmp_diff_equation} to Eq.~\ref{eq:dmp_scaled_diff_equation}, and $Q_K$ denotes the DMP-exaggerated motion. This gives $K$ an interpretation as the exaggeration intensity in which $K=1$ reconstructs the original neutral motion, while $K>1$ amplifies the motion shape. 
            % In Appendix~\ref{app:theoretical_analysis}, we provide a theoretical analysis of our DMP-exaggeration formulation, focusing on two practical considerations: key-pose spacing and forcing-basis capacity.
        }

        % \vspace{-5pt}
        
        \subsubsection{Exaggeration Formulation for Guiding Diffusion and Flow Matching}
        \label{sec:tf_formulation_diffusion}
        
            % \vspace{-5pt}

            The DMP provides a controllable mechanism for exaggeration but has no knowledge of physically plausible character motion. Therefore, naively exaggerating using DMP (Eq.~\ref{eq:dmp_exag_operator}), especially at large $K$, can introduce artifacts such as invalid body poses and unrealistic movements. To address this issue, we leverage a pre-trained motion generation model to iteratively refine the DMP-exaggerated motion, keeping it close to the physically plausible motion manifold learned during pre-training. This is possible because motion foundation models learn a strong prior over realistic motion capture data and implicitly capture the physical properties of character motion during inference. However, they do not explicitly model the animation principle of exaggeration. Therefore, we combine DMPs with a pre-trained motion generation model to generate motions that are both expressive/exaggerated and physically plausible. In this way, the DMP determines the exaggerated trajectory (Eq.~\ref{eq:dmp_exag_operator}), while the pre-trained model iteratively regularizes it using its learned physical motion prior. We refer to this approach as \textit{training-free (TF)}, or inference-time exaggeration, since it modifies only the inference process and requires no model training. We consider both diffusion \citep{Kimodo, MDM} and flow-matching \citep{HY-Motion} motion generation models. The main paper focuses on the diffusion formulation. The flow-matching formulation is provided in Appendix~\ref{app:remarks_training_free_exaggeration}.
        
            % \vspace{-4pt}
    
            \textbf{Integrating DMP-based exaggerated motion ($Q_K$) into diffusion denoising process.} A motion diffusion model defines a forward process that gradually corrupts clean motion into Gaussian noise and a learned reverse process that reconstructs it. Following the clean-signal (\(x_0\)) parameterization of MDM \citep{MDM}, we denote the noisy motion latent at diffusion step \(s\) by \(z_s\) and the model's clean-motion prediction by $\hat{Q}^{(s)} = \mathcal{G}_{\theta}(z_s,s,c).$ Sampling starts from \(z_S\sim\mathcal{N}(0,I)\) and iteratively denoises the sample to obtain the final clean motion at \(s=0\). Our inference-time approach modifies this sampling process while keeping \(\theta\) unchanged. Given a condition $c$, we first use the frozen model $\mathcal{G}_{\theta}$ to generate a neutral motion $Q_{\mathrm{neut}}$. Then, we use Eq.~\ref{eq:dmp_exag_operator} to obtain the DMP-based exaggerated motion $Q_K$. Then, we repeat this reverse diffusion process using the same initial noise that generated $Q_{\mathrm{neut}}$. At each denoising step $s$, the frozen model predicts a cleaner motion, $\hat{Q}^{(s)}=\mathcal{G}_{\theta}(z_s,s,c)$, by removing noise from the output of the previous denoising step $s-1$. To exaggerate this motion, before performing the next diffusion update, we move the prediction $\hat{Q}^{(s)}$ toward the DMP-exaggerated target $Q_K$. We repeat this process for $S$ diffusion denoising steps. This process can be written as below:
            
            \vspace{-10pt}
            
            \begin{equation}
                \label{eq:guide}
                \widetilde{Q}^{(s)}
                =
                \hat{Q}^{(s)}
                +
                \omega_s\big(Q_K-\hat{Q}^{(s)}\big),
                \qquad
                z_{s-1}=\mathrm{DDIM}\big(z_s,\widetilde{Q}^{(s)}\big).
            \end{equation}
            
            \vspace{-5pt}
            
            where $\omega_s$ denotes the exaggeration guidance strength at diffusion step $s$. We parameterize it as $\omega_s=\omega\,\gamma(s)$ where $\omega$ is the overall guidance scale and $\gamma(s)\in[0,1]$ controls how strongly the exaggeration target is applied throughout the denoising trajectory. We use a heuristic schedule where the guidance remains weak during the earliest, highly noisy diffusion steps, and instead becomes stronger during the intermediate part of the generation process, and decreases again near the final denoising steps. This design was chosen to avoid forcing unreliable early predictions toward the target and gives the model additional freedom in the final steps to produce a clean and plausible motion. Eq.~\ref{eq:guide} balances two objectives: the difference $Q_K-\hat{Q}^{(s)}$ moves the current prediction toward the DMP-exaggerated motion, while passing the guided state through $\mathcal{G}_{\theta}$ at the next denoising step reintroduces the learned motion prior. In this way, the iterative refinement process corrects implausible poses introduced by DMP-based exaggeration (Eq.~\ref{eq:dmp_exag_operator}).
    }

\vspace{-5pt}

\section{Experiments and Results}
\label{Experiments_Results}
{    
    \vspace{-5pt}

    \begin{figure}[t]
        \vspace{-10pt}
        \centering
        \includegraphics[width=1\linewidth]{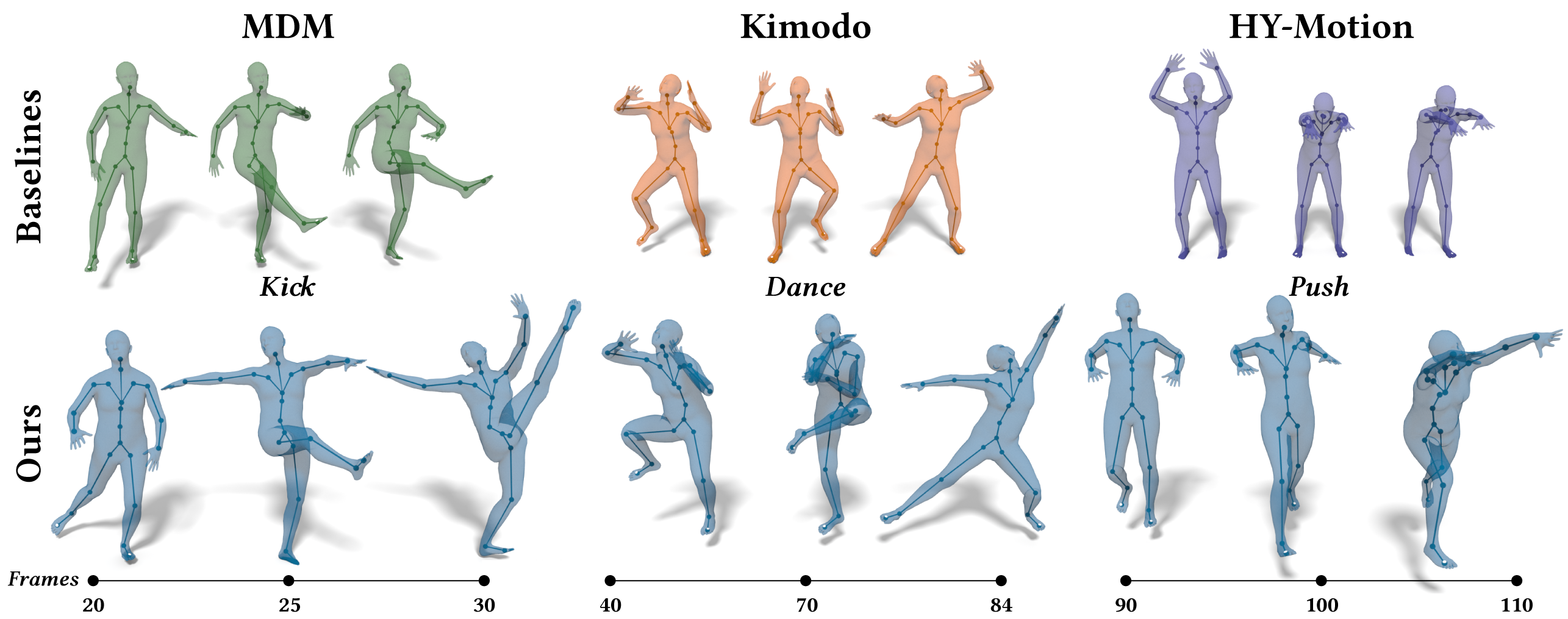}
        \vspace{-15pt}
        \caption{\small \textbf{Qualitative comparison of motion exaggeration.} For each action, the top row shows the motion generated by the pre-trained model, while the bottom row shows the corresponding motion generated with our exaggeration method. Across MDM (green), KIMODO (orange), and HY-Motion (purple), our method (blue) consistently produces more expressive and exaggerated movements while preserving the underlying action.}
        \vspace{-5pt}
        \label{fig:main_result}
    \end{figure}

    We evaluate our training-time (SFT), inference-time (TF), and combined (SFT+TF) exaggeration approaches across a diverse collection of character motions. We apply these approaches to three pre-trained text-to-motion backbones including KIMODO~\citep{Kimodo}, HY-Motion-1.0~\citep{HY-Motion}, and MDM~\citep{MDM}. We evaluate the resulting motions through qualitative comparisons, quantitative metrics, ablation studies, and a perceptual user study. Additional experimental results are provided in Appendix~\ref{app:additional_experiments_results}.

    % \vspace{-7pt}
        
    \subsection{Qualitative Evaluation}
    \label{qualitative_evaluation}
    % \vspace{-5pt}
    
    Figs.~\ref{fig:teaser} and~\ref{fig:main_result} present representative qualitative comparisons for three diverse actions including kick, dance, and push across three baselines: MDM, KIMODO, and HY-Motion. Compared to the baselines, our approaches produce more expressive and exaggerated motions while preserving the underlying actions. Additional results covering more diverse actions, and exaggeration approaches (TF, SFT, and SFT+TF) are provided in Figs.~\ref{fig:result_sft+tf}--\ref{fig:result_controllable_exaggeration_tf} in Appendix~\ref{app:additional_experiments_results}. We also encourage readers to view the supplemental video for additional qualitative comparisons with the baselines and for a more complete assessment of the resulting motions and their temporal progression.

    \vspace{-5pt}
    
    \subsection{Quantitative Evaluation}
    \label{sec:quantitative_evaluation}
    
        % \vspace{-5pt}

        We evaluate the proposed motion exaggeration methods along two evaluation axes: \textit{(i) exaggeration ability}, where we assess how well the method can produce exaggerated and expressive motions, and \textit{(ii) action preservation}, where we assess how well the method can preserve the intent and physical properties of the motion. An ideal expressive motion generation method should achieve strong exaggeration while preserving the underlying action and motion quality. However, stronger exaggeration naturally shifts the generated motions away from the original training distribution, which can affect distribution-based metrics such as FID even when the action remains recognizable and the motion physically plausible. Our results in Table~\ref{tab:quantitative_results} show that the proposed methods remain competitive on standard action-preservation and quality metrics while consistently achieving stronger exaggeration, with the combined SFT+TF approach providing the strongest overall exaggeration performance across the evaluated models.
        
        % \vspace{-4pt}

        \noindent \textbf{Motion Exaggeration Evaluation.}
        To the best of our knowledge, this is the first work to incorporate the \textit{Exaggeration} principle of animation into motion generation models and to introduce a dedicated evaluation suite for quantifying exaggeration in generated motions. Existing motion-generation benchmarks mainly evaluate text-motion alignment and generation quality, but do not assess whether a model produces the requested degree of exaggeration. To address this gap, we propose two complementary groups of metrics: \emph{(i) classifier-based metrics}, which use an independent exaggeration-level classifier trained on \datasetname~to determine whether a generated motion exhibits the requested intensity level, and \emph{(ii) classifier-free metrics}, which directly measure physical motion intensity from skeleton joint trajectories without relying on a learned classifier. The classifier-based group includes \(\mathrm{ELRA}@k\), \(k\in\{1,2,3\}\), which measures how often the requested exaggeration level appears among the classifier's top-\(k\) predictions, and \(\rho_{\mathrm{clf}}\), the Spearman rank correlation~\citep{Spearman1904} between the requested and predicted levels, which measures whether the predictions follow the requested progression from neutral to extreme exaggeration. For the classifier-free metrics, we use root-relative range of motion (\textit{ROM}) as the physical-intensity score, reducing the influence of global character translation. We report \(\rho_{\mathrm{phys}}\), the action-averaged Spearman rank correlation between the requested level and physical intensity, and \(\Delta_{\mathrm{phys}}\), the standardized intensity difference between neutral and extremely exaggerated motions. These two metrics, \(\rho_{\mathrm{phys}}\) and \(\Delta_{\mathrm{phys}}\), capture the consistency and magnitude of physical amplification, respectively. Also, following ~\citep{AnimationBench}, we report \textit{Novelty} (\textit{Nov.}), which measures the feature-space deviation between a highest-level exaggerated motion and its paired neutral reference using the pretrained HumanML3D motion encoder. Detailed definitions of all metrics are provided in Appendix~\ref{app:remarks_exaggeration_evaluation_suit}.
        
        % \vspace{-4pt}

        \begin{table*}[t] 
            \vspace{-5pt}
            \centering 
            \caption{\small \textbf{Quantitative evaluation of exaggeration and action preservation.} We report results for HY-Motion (1B), MDM, and KIMODO under the baseline, DMP, TF, SFT, and combined SFT+TF settings. The Exaggeration metrics measure exaggeration control and physical intensity, while Action Preservation metrics evaluate text-motion alignment and preservation of the learned motion manifold. Quality reports foot-skate artifacts. $\uparrow$ and $\downarrow$ indicate higher and lower values are better, respectively. Bold values highlight the best result within each model and metric, while underlined values indicate the second-best result.}
            \vspace{-5pt}
            \label{tab:quantitative_results} 
            \resizebox{\textwidth}{!}{% 
            \begin{tabular}{llcccccccccc} 
            \toprule 
             
            & & \multicolumn{5}{c}{\textbf{Exaggeration}} 
            & \multicolumn{4}{c}{\textbf{Action Preservation}} 
            & \multicolumn{1}{c}{\textbf{Quality}} \\ 
             
            \cmidrule(lr){3-7} 
            \cmidrule(lr){8-11} 
            \cmidrule(lr){12-12} 
             
            \textbf{Model} & \textbf{Method} 
            & $\mathrm{ELRA}@1\uparrow$ 
            & $\rho_{\mathrm{clf}}\uparrow$ 
            & $\rho_{\mathrm{phys}}\uparrow$ 
            & $\Delta_{\mathrm{phys}}\uparrow$ 
            & Nov.$\uparrow$ 
            & R@3$\uparrow$ 
            & FID$\downarrow$ 
            & Div$\uparrow$ 
            & MMDist$\downarrow$ 
            & Skate$\downarrow$ \\ 
             
            \midrule 
            
            \multirow{1}{*}{\textbf{MDM}} 
                & Baseline 
                & \underline{0.200} & 0.000 & 0.026 & 0.175 & 0.117 
                & \underline{0.7031} 
                & \textbf{0.8388} & \textbf{9.5722} & \textbf{3.6133} & \textbf{0.070} \\ 
             
                & \hspace{0.5em} + DMP 
                & \underline{0.200} & 0.000 & 0.358 & 0.953 & 0.130 
                & 0.5625 & 20.9116 & 6.8088 & 5.2530 & 0.134 \\ 
             
                & \hspace{0.5em} + TF 
                & \underline{0.200} & 0.000 & 0.252 & 0.552 & 0.023 
                & \textbf{0.7070} & 5.7547 & 8.3935 & \underline{3.9870} & 0.091 \\ 
             
                & \hspace{0.5em} + SFT 
                & \textbf{0.408} & \underline{0.635} & \underline{0.399} & \underline{1.173} & \underline{0.168} 
                & 0.5547 & \underline{1.6349} & \underline{8.8127} & 4.6498 & \underline{0.073} \\ 
             
                & \hspace{0.5em} + SFT+TF 
                & \textbf{0.408} & \textbf{0.643} & \textbf{0.545} & \textbf{1.610} & \textbf{0.276} 
                & 0.5273 & 5.8575 & 7.7830 & 5.0234 & 0.085 \\ 
             
            \midrule 
             
            \multirow{1}{*}{\textbf{HY-Motion-1B}} 
                & Baseline 
                & 0.354 & 0.354 & 0.391 & 0.902 & 0.135 
                & \textbf{0.7422} 
                & \textbf{2.0825} & \textbf{8.9325} & \textbf{3.4851} & \textbf{0.029} \\ 
             
                & \hspace{0.5em} + DMP 
                & 0.200 & -0.042 & 0.516 & 1.318 & 0.167 
                & 0.4023 & 30.8515 & 4.7568 & 6.0145 & 0.061 \\ 
             
                & \hspace{0.5em} + TF 
                & 0.200 & 0.007 & 0.424 & 1.017 & 0.082 
                & 0.5898 & 17.0509 & 6.4124 & 4.8573 & 0.057 \\ 
             
                & \hspace{0.5em} + SFT 
                & \textbf{0.938} & \textbf{0.985} & \underline{0.556} & \underline{1.622} & \underline{0.222} 
                & \underline{0.7383} & \underline{2.2297} & \underline{8.4439} & \underline{3.6413} & \underline{0.040} \\ 
             
                & \hspace{0.5em} + SFT+TF 
                & \underline{0.900} & \underline{0.973} & \textbf{0.697} & \textbf{2.003} & \textbf{0.301} 
                & 0.5430 & 19.7208 & 5.8073 & 5.1643 & 0.063 \\ 
             
            \midrule 
             
            \multirow{1}{*}{\textbf{KIMODO}} 
                & Baseline 
                & 0.223 & -0.051 & 0.137 & 0.281 & 0.080 
                & \textbf{0.6562} 
                & \underline{2.0376} & \underline{8.5198} & \textbf{3.8190} & \textbf{0.025} \\ 
             
                & \hspace{0.5em} + DMP 
                & 0.208 & 0.053 & 0.374 & 1.086 & 0.144 
                & 0.3516 & 38.1156 & 5.7478 & 6.7370 & 0.069 \\ 
             
                & \hspace{0.5em} + TF 
                & 0.208 & 0.039 & 0.287 & 0.801 & 0.081 
                & 0.4961 & 20.0688 & 6.8045 & 5.3506 & 0.036 \\ 
             
                & \hspace{0.5em} + SFT 
                & \underline{0.515} & \underline{0.814} & \underline{0.587} & \underline{1.610} & \underline{0.183} 
                & \underline{0.6289} & \textbf{1.6934} & \textbf{8.9660} & \underline{3.9654} & \underline{0.027} \\ 
             
                & \hspace{0.5em} + SFT+TF 
                & \textbf{0.531} & \textbf{0.831} & \textbf{0.729} & \textbf{2.015} & \textbf{0.306} 
                & 0.4648 & 18.9754 & 7.2581 & 5.5249 & 0.030 \\ 
             
            \bottomrule 
            \end{tabular}}
            \vspace{-5pt}
        \end{table*}

        \noindent \textbf{Action-Preservation Evaluation.} We evaluate action preservation and motion quality using the HumanML3D \cite{Guo2022HumanML3D} evaluation protocol. We report \textit{R-precision}@\(k\) (\(k\in{1,2,3}\)) and \textit{MMDist} for text-motion alignment, \textit{FID} and \textit{Diversity} for motion-distribution quality, and \textit{Foot-Skate} for physical plausibility. Lower values indicate better performance for \textit{MMDist}, \textit{FID}, and \textit{Foot-Skate}, while higher values are better for the remaining metrics.

        % \vspace{-4pt}

        \textbf{Ablation Studies.} To analyze the design choices in our proposed method, we perform a set of controlled ablations and summarize the key findings below: 

        \begin{wrapfigure}[33]{r}{0.4\textwidth}
            \vspace{-40pt}
            \centering
        
            % --- (a) LoRA -------------------------------
            \resizebox{1\linewidth}{!}{%
            \includegraphics{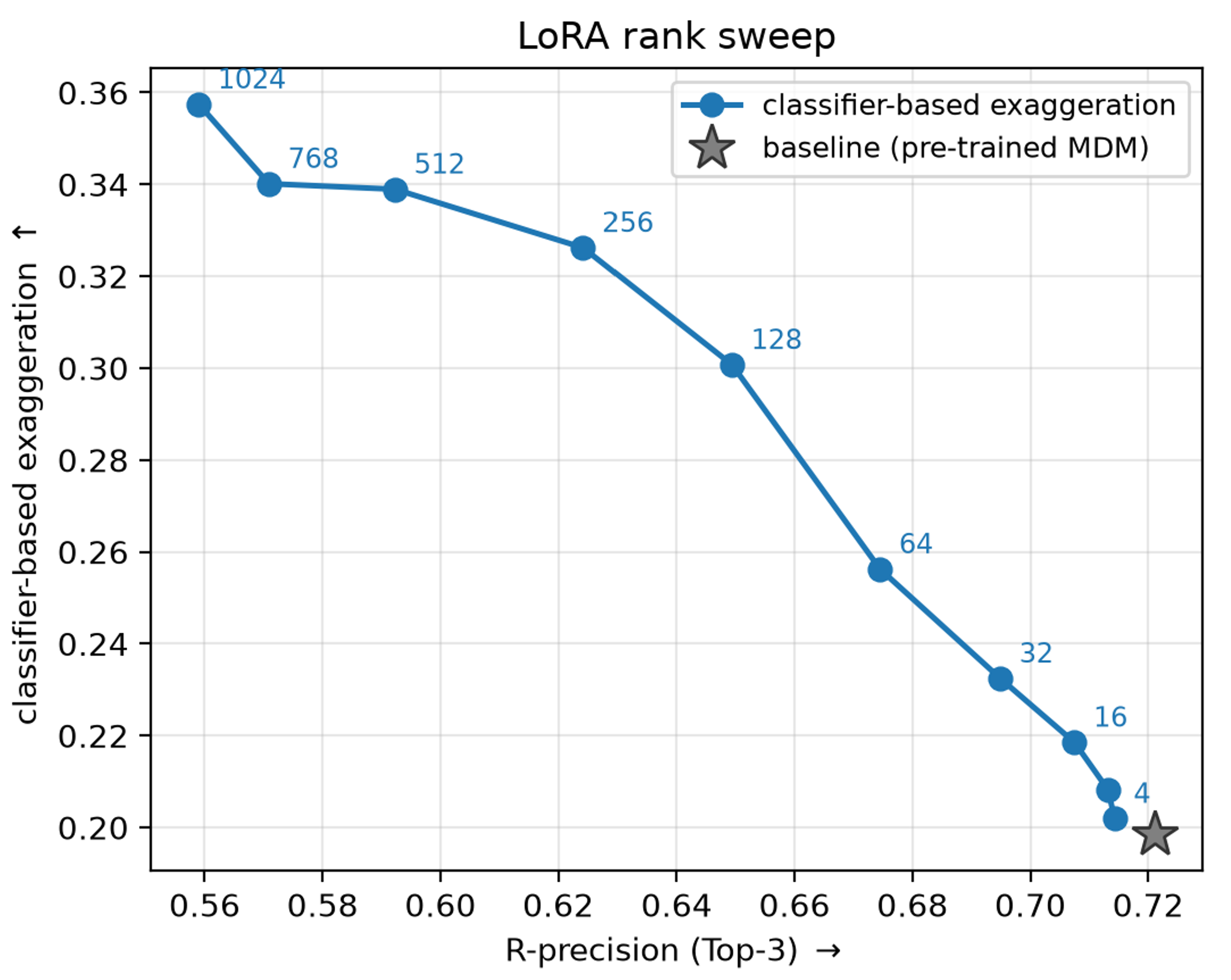}%
            }
            \vspace{-20pt}
            \caption*{\small (a) Ablation on $r$}
            
            \vspace{5pt}
        
            % --- (b) Lambda -----------------------------
            \resizebox{1\linewidth}{!}{%
            \includegraphics{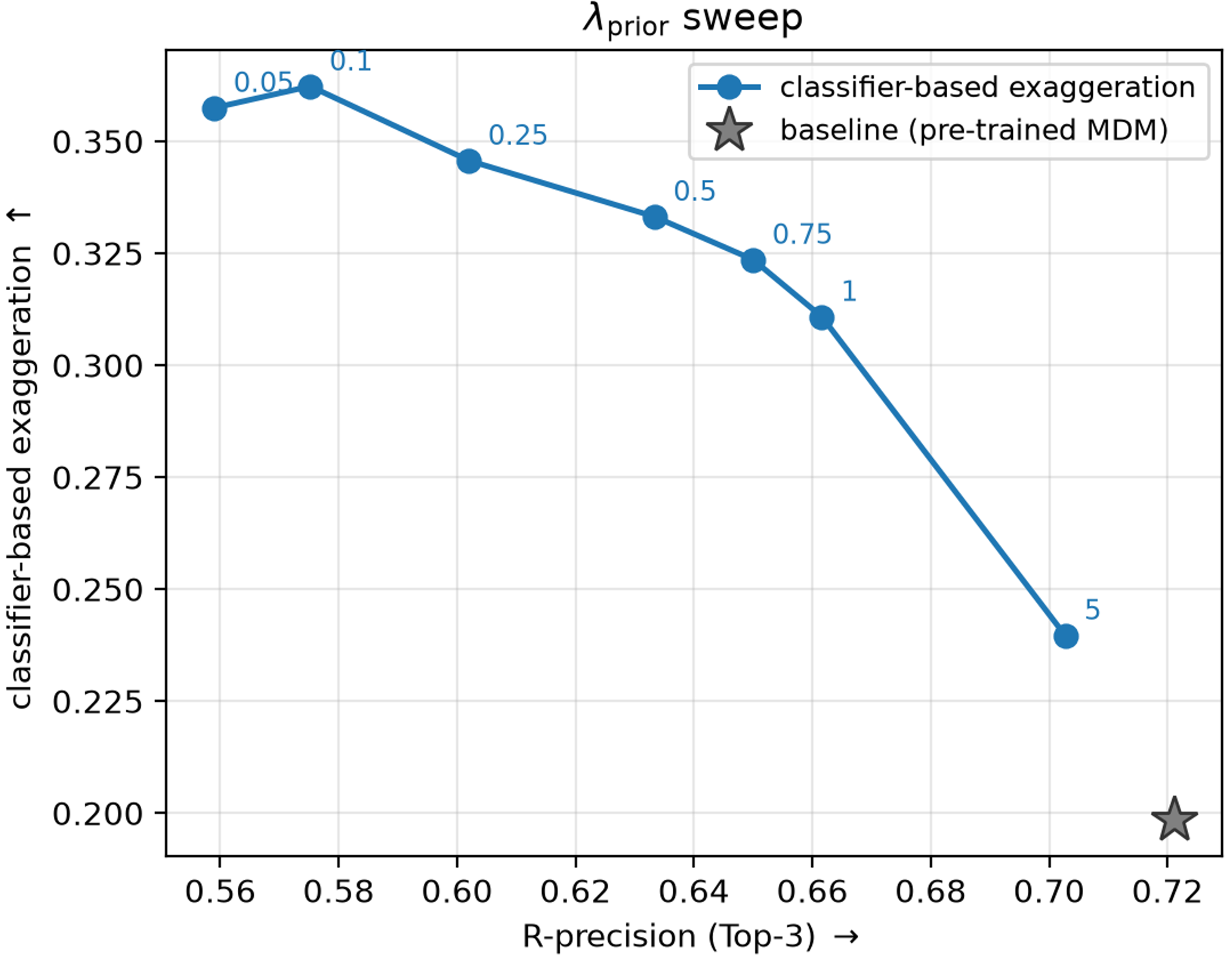}%
            }
            \vspace{-20pt}
            \caption*{\small (b) Ablation on $\lambda$}
            % \vspace{-5pt}
        
            \caption{\small \textbf{Ablation study on LoRA rank and prior-preservation weight.} We evaluate the trade-off between classifier-based exaggeration ($\mathrm{ELRA}@1$) and prompt following by varying  (a) the LoRA rank $r$ and  (b) prior-preservation weight $\lambda$ for MDM. Higher $\mathrm{ELRA}@1$ indicates stronger exaggeration control, while higher R-precision indicates better prompt adherence. The pre-trained MDM baseline is shown by the star.}
            \label{fig:mdm_ablation}                
        \end{wrapfigure}

        \textit{(i) Training strategy.} We compare two SFT strategies: full fine-tuning and the proposed LoRA+Prior-preservation fine-tuning recipe. Quantitative results in Tables~\ref{tab:mdm_sft_ablation}-\ref{tab:hy1b_sft_ablation} in Appendix~\ref{app:additional_experiments_results} show that full fine-tuning produces more exaggerated motions but worse action preservation. In contrast, LoRA+Prior-preservation achieves a better trade-off, maintaining strong exaggeration while better preserving the action intent and type. 
        
        \textit{(ii) LoRA ranks and prior-preservation weights.} We ablate the proposed LoRA+Prior-preservation fine-tuning setting by training identical models with different values of LoRA rank \(r\) and prior-preservation weight \(\lambda\). Fig.~\ref{fig:mdm_ablation} shows that increasing the LoRA rank \(r\) improves exaggeration control (higher $\mathrm{ELRA}@1$) but reduces text adherence (lower R@3), while increasing the prior-preservation weight \(\lambda\) has the opposite effect. Therefore, we choose an operating point with \(r\in\{64,128\}\) and \(\lambda\in\{0.1,0.25\}\) depending on the base model to achieve a good trade-off. Tables~\ref{tab:mdm_sft_ablation}-\ref{tab:kimodo_sft_ablation} in Appendix~\ref{app:additional_experiments_results} present the full ablation on $r$ and $\lambda$ for MDM, HY-Motion, and KIMODO. 
        
        \textit{(iii) Exaggeration components analysis.} We replace the exaggeration components proposed in this study. DMP, TF, SFT, and their composition (SFT+TF) are all designed to exaggerate character motion. As shown in Table~\ref{tab:quantitative_results}, across all three motion models, DMP consistently increases physical exaggeration but substantially reduces action preservation and motion quality. Adding TF recovers action preservation and physical plausibility, demonstrating the value of the pre-trained motion prior for constraining DMP-exaggerated motion. SFT provides stronger control over the requested exaggeration level while better preserving the original motion quality than DMP and SFT+TF. Finally, SFT+TF achieves the strongest physical exaggeration across the models, but at the cost of reduced action preservation and increased motion artifacts. 

        \begin{wraptable}{r}{0.4\columnwidth}
            % \vspace{-5pt}
            \centering
            \caption{\small \textbf{User-study results.} \emph{Exaggeration} reports how often participants selected the highest requested level as the most exaggerated, while \emph{Action Preservation} reports the percentage of positive preservation ratings. 
            % ``--'' indicates that the backbone was not evaluated for action preservation.
            }
            \label{tab:user_study}
            \vspace{-4pt}
            \scriptsize
            \setlength{\tabcolsep}{8pt}
            \renewcommand{\arraystretch}{1}
            \begin{tabular}{lcc}
                \toprule
                \textbf{Backbone}
                & \textbf{Exaggeration} & \shortstack{\textbf{Action}\\\textbf{Preservation}} \\
                \midrule
                KIMODO    & 86.7\% & 82.0\% \\
                HY-Motion & 78.9\% & 50.0\% \\
                MDM       & 70.0\% & --   \\
                \midrule
                \textbf{Overall}
                          & \textbf{81.8\%} & \textbf{76.7\%} \\
                \bottomrule
            \end{tabular}
            \vspace{-2pt}
        \end{wraptable}
        
        Overall, the results demonstrate the expected exaggeration-preservation trade-off: DMP provides direct but unconstrained amplification, TF improves plausibility, SFT offers stronger exaggeration-level control, and SFT+TF maximizes physical exaggeration at the cost of greater deviation from the original action. The complete version of Table~\ref{tab:quantitative_results} is presented in Table~\ref{tab:ablation_exaggeration_components} in Appendix~\ref{app:additional_experiments_results}.

        \textbf{User Study.} To perceptually assess the exaggeration capability of our method and its ability to preserve the underlying action, we conducted a user study with 20 participants with diverse backgrounds. For the \emph{exaggeration} evaluation, participants viewed neutral and progressively exaggerated motions side-by-side and selected the most exaggerated and most neutral motions, resulting in 10 questions per participant. We report the \emph{top-1 selection rate}, measuring how often the highest requested exaggeration level was selected as the most exaggerated, together with the neutral-reference selection rate. For the \emph{action-preservation} evaluation, participants viewed a source motion and its exaggerated counterpart and judged whether the counterpart preserved the source action, resulting in 6 additional questions. 
        
        Across the 16 evaluations (320 samples), the highest requested exaggeration level was selected as the most exaggerated in 81.8\% of valid trials, while the exaggerated motions were judged to preserve the source action in 76.7\% of cases. As summarized in Table~\ref{tab:user_study}, these results show that our approach produces clearly perceptible motion exaggeration while preserving the underlying action in most evaluations. The complete study protocol, representative questionnaire examples, and detailed results are provided in Appendix~\ref{app:additional_experiments_results}, Fig.~\ref{fig:user_study_questionnaire}, and Table~\ref{tab:user_study_full}.
}

% \vspace{-5pt}

\section{Conclusions}
\label{Conclusions}
{
    % \vspace{-5pt}
    This study introduces a framework for incorporating the \textit{Exaggeration} principle of animation into modern motion generative foundation models through training-time and inference-time approaches. Training-time exaggeration uses supervised fine-tuning on our graded exaggeration dataset, while inference-time exaggeration formulates exaggeration with dynamic movement primitives and uses the resulting signal to guide diffusion and flow-matching models without additional training. Qualitative and quantitative evaluations across multiple motion generation models demonstrate that our approaches generate more expressive motions while preserving the underlying action. 
    
    Our work highlights the potential of incorporating established character animation principles beyond physical realism into motion generation. We hope this work encourages further research toward integrating additional animation principles into motion foundation models. One limitation of our current formulation is that the appropriate exaggeration scale is motion dependent. This means applying a very large \(K\) across diverse motions can occasionally produce jittery or atypical outputs. Future work could investigate reinforcement-learning-based post-training to improve motion plausibility while preserving the intended expressiveness.

    %%%%%%%%%%%%%%%% Add the future work -> RL post-training %%%%%%%%%%%%%%%%%%%%%%%
}
    

% }

\subsubsection*{Acknowledgments}
{
    This research was supported by Autodesk Research and the Autodesk AI Lab. We thank Noa Kaplan, Larasika Nadela, and Lucie Taglienti for their contributions to the dataset, and Marc Chu, Robert Helms, Helen Lu, Evan Atherton, Shu Ishida, and Jenmy Zhang for valuable discussions and support throughout this work.
}

\clearpage 

% Recompile several times to resolve all references
\bibliography{iclr2027_conference}
\bibliographystyle{iclr2027_conference}

\clearpage

\appendix

\section{Remarks on 12 Principles of Animation}
\label{app:Remarks_12_AnimationPrinciples}
{
    \vspace{-5pt}
    
    The 12 Principles of Animation were introduced by Disney animators as a set of principles for designing and producing character animation \citep{AnimationPrinciples}. These principles provide a practical framework for creating animation that is not only physically understandable, but also expressive and engaging to the audience. Although originally developed in the context of traditional hand-drawn animation, these principles remain relevant to computer animation and have been adapted to 3D character animation \citep{3DComputerAnimation}. Below, we briefly describe the 12 principles and their role in character animation.    
    
    \textbf{1. Squash and Stretch} describes the deformation of a character as it moves. Squashing and stretching can be used to communicate weight, flexibility, and the forces acting on a character. For example, a character may compress when landing and extend during a fast movement.

    \vspace{-3pt}
    
    \textbf{2. Anticipation} prepares the audience for an upcoming action. A character typically performs a preparatory movement before executing the main action, such as bending the body before jumping or pulling an arm backward before throwing an object. Anticipation makes the subsequent action easier to perceive.

    \vspace{-3pt}
    
    \textbf{3. Staging} concerns how an action, pose, or idea is presented so that its intention is clear to the audience. It involves the arrangement of the character, its pose, movement, and the surrounding scene to direct the viewer's attention toward the most important aspect of the action.    

    \vspace{-3pt}
    
    \textbf{4. Straight Ahead Action and Pose-to-Pose} describe two approaches to constructing an animated sequence. In straight-ahead animation, motion is created sequentially from one frame to the next, which can produce spontaneous and fluid motion. In pose-to-pose animation, important key poses are first established and the intermediate poses are subsequently generated between them.

    \vspace{-3pt}
    
    \textbf{5. Follow Through and Overlapping Action} describe the fact that different parts of a character do not necessarily stop or change direction simultaneously. Parts such as the arms, legs, head, clothing, or other secondary elements may continue moving after the main body action has changed. This asynchronous movement helps produce more natural and dynamic character motion.

    \vspace{-3pt}
    
    \textbf{6. Slow In and Slow Out} describes the gradual acceleration and deceleration of a movement. Rather than moving at a constant speed between two poses, a character often starts slowly, moves faster through the middle of the action, and slows down when approaching the next pose. This principle is commonly used to make motion feel smoother and more natural.

    \vspace{-3pt}
    
    \textbf{7. Arcs} refers to the tendency of natural character movements to follow curved trajectories rather than perfectly straight paths. For example, the movement of an arm during a swing or the motion of a head during a turn generally follows an arc. Designing motion along appropriate arcs can make character movement appear more natural and fluid.

    \vspace{-3pt}
    
    \textbf{8. Secondary Action} refers to an additional movement that supports and reinforces the main action without distracting from it. For example, while a character walks, arm movements, head motion, or changes in facial expression can provide additional information about the character's state and intention. Secondary actions can increase the expressiveness of a primary action.

    \vspace{-3pt}
    
    \textbf{9. Timing} controls the temporal structure of an action, including how long a movement takes and how motion is distributed throughout the sequence. Different timing can communicate different properties, such as weight, speed, strength, emotion, or personality. Consequently, the same action can convey a different meaning or feeling when performed with different timing.

    \vspace{-3pt}
    
    \textbf{10. Exaggeration} emphasizes or amplifies the characteristics of an action to make its intention, expression, or emotional quality clearer and more engaging. Importantly, exaggeration does not necessarily imply changing the underlying action. Instead, an animator can amplify the shape, timing, pose, or movement characteristics of an existing action while preserving its identity. For example, a punch can be exaggerated by increasing the extent or speed of its movement while still being recognized as the same punch action. The degree of exaggeration can range from subtle amplification to highly stylized motion, depending on the desired artistic effect.
    
    \vspace{-3pt}
    
    \textbf{11. Solid Drawing} concerns the ability to construct and animate characters while maintaining a consistent understanding of their three-dimensional form, proportions, balance, weight, and perspective. Although originally formulated for hand-drawn animation, the principle emphasizes maintaining a coherent representation of the character as it moves through space.

    \vspace{-3pt}
    
    \textbf{12. Appeal} refers to the visual and behavioral qualities that make a character interesting and engaging to the audience. An appealing character does not necessarily have to be attractive. Rather, its design, expressions, and movements can also create a clear and compelling presence. In character animation, appeal can arise from both the character's appearance and the way the character moves.
}

\section{Remarks on the Exaggeration Evaluation Suite}
\label{app:remarks_exaggeration_evaluation_suit}
{
    \vspace{-5pt}

    To evaluate whether a generated motion exhibits the requested degree of exaggeration, we propose two complementary groups of metrics: \emph{classifier-based} and \emph{classifier-free}. Let \(N\) denote the number of evaluated motions, and let \(l_i\in\{0,1,2,3,4\}\) denote the requested exaggeration level for motion \(i\), ranging from neutral (\(l_i=0\)) to extremely exaggerated (\(l_i=4\)).

    \vspace{-5pt}

    \subsection{Classifier-Based Exaggeration Metrics.}
    {
        \vspace{-5pt}

        \textbf{Exaggeration-Level Classifier.} We first evaluate whether a generated motion is recognized as having the requested exaggeration level. To this end, we train a motion classifier to predict the five exaggeration levels \(l\in\{0,1,2,3,4\}\), corresponding to neutral, slightly, moderately, very, and extremely exaggerated motion. The classifiers are trained from scratch on real labeled motions from \datasetname~and do not receive action captions or generated motions as input. Once trained, their parameters remain frozen throughout evaluation. For MDM \citep{MDM}, the classifier operates directly on the normalized \(263\)-dimensional HumanML3D representation \citep{Guo2022HumanML3D} used by the generative model. For HY-Motion \citep{HY-Motion} and KIMODO \citep{Kimodo}, we instead use the \(3\)D positions of the \(22\) body joints after subtracting the root position at every frame, resulting in a \(66\)-dimensional translation-invariant representation. HY-Motion joint positions are obtained from the joint-position block of its native \(201\)-dimensional representation, and KIMODO joint positions are decoded from its native \(273\)-dimensional representation using its forward-kinematics decoder. Separate HY-Motion and KIMODO classifiers are trained because the motions are produced through different native representations and decoding procedures. The HY-Motion-0.4B and HY-Motion-1B models share the same classifier because they use the same motion representation.
        
        All three classifiers use the same architecture. Given an input motion \(\mathbf{X}\in\mathbb{R}^{C\times T_c}\), two one-dimensional convolutional layers with \(256\) channels, kernel size \(3\), and padding \(1\) are applied along the temporal dimension. Each convolution is followed by batch normalization and a ReLU activation. Then, adaptive temporal average pooling produces a single \(256\)-dimensional motion descriptor, which is passed through dropout with probability \(0.3\) and a linear layer producing five level logits. We use \(T_c=100\) frames for MDM at \(20\) FPS and \(T_c=150\) frames for HY-Motion and KIMODO at \(30\) FPS, corresponding to a \(5\)-second input window for every classifier. Longer training clips are randomly cropped. Each classifier is trained using five-class cross-entropy loss and the Adam optimizer \citep{Adam} with a learning rate of \(5\times10^{-4}\), weight decay \(10^{-4}\), and batch size \(64\). Training runs for at most \(300\) epochs, with early stopping after \(40\) epochs without improvement in exact validation accuracy. 
        
        During evaluation, every generated motion is converted to the corresponding classifier input representation, standardized using the classifier's stored statistics, and processed using the deterministic evaluation crop. Let \(\mathbf{p}_i\in\mathbb{R}^{5}\) denote the resulting classifier logits for motion \(i\), and let \(\hat l_i=\arg\max_m p_{i,m}\) denote its predicted exaggeration level. The same frozen classifier checkpoint is used to evaluate every method and ablation associated with a given backbone. Because the classifiers operate in different representation spaces, their absolute scores are intended primarily for comparisons among methods evaluated on the same backbone.

        \textbf{Exaggeration-Level Recognition Accuracy.} Using the frozen exaggeration-level classifier described above, we define $\mathrm{ELRA}@1$ as
        
        \vspace{-10pt}
        
        \begin{equation}
            \mathrm{ELRA}@1
            =
            \frac{1}{N}
            \sum_{i=1}^{N}
            \mathbf{1}\!\left[
                l_i\in\mathrm{Top}_1(\mathbf{p}_i)
            \right],
        \end{equation}
        
        \vspace{-5pt}
        
        where \(\mathrm{Top}_1(\mathbf{p}_i)\) denotes the set of exaggeration-level indices corresponding to the classifier's largest logits for motion \(i\), and \(\mathbf{1}[\cdot]\) is the indicator function. Therefore, \(\mathrm{ELRA}@1\) measures the fraction of generated motions for which the requested exaggeration level is included among the classifier's top-1 predictions. Higher values indicate that generated motions more consistently exhibit the requested exaggeration level according to the frozen classifier.

        \textbf{Exaggeration-Level Rank Correlation.}
        Because the exaggeration levels are ordinal, we additionally measure whether the classifier's predictions increase consistently as the requested level increases. We use Spearman's rank correlation coefficient~\citep{Spearman1904}, which, for two sequences \(\mathbf{u}\) and \(\mathbf{v}\), is defined as
        
        \vspace{-10pt}
        
        \begin{equation}
            \rho_{\mathrm{S}}(\mathbf{u},\mathbf{v})
            =
            \mathrm{Corr}_{\mathrm{P}}\!\left(
                R(\mathbf{u}),R(\mathbf{v})
            \right),
        \end{equation}
        
        \vspace{-5pt}
        
        where \(R(\cdot)\) converts the observations to ranks, assigning average ranks to tied values, and \(\mathrm{Corr}_{\mathrm{P}}\) denotes Pearson correlation between the resulting ranks. Spearman correlation lies in \([-1,1]\), where \(1\) indicates a perfectly monotonic increase, \(0\) indicates no monotonic association, and \(-1\) indicates a perfectly monotonic decrease. Using the classifier's top-1 prediction \(\hat l_i\), we define the classifier-based rank correlation as
        
        \vspace{-10pt}
        
        \begin{equation}
            \rho_{\mathrm{clf}}
            =
            \rho_{\mathrm{S}}\!\left(
                \{l_i\}_{i=1}^{N},
                \{\hat l_i\}_{i=1}^{N}
            \right).
        \end{equation}
        
        \vspace{-5pt}
        
        While \(\mathrm{ELRA}@1\) measures per-motion agreement between the requested level and the classifier's predictions, \(\rho_{\mathrm{clf}}\) measures whether the predicted levels follow the overall ordering of the requested exaggeration progression. Therefore, the two metrics are complementary: a model may preserve the correct ordering of exaggeration levels even when some individual levels are not classified exactly. A higher \(\rho_{\mathrm{clf}}\) indicates more consistent control over the progression from neutral to extreme exaggeration.
    }

    \vspace{-5pt}
    
    \subsection{Classifier-Free Exaggeration Metrics.}
    {
        \vspace{-5pt}
        
        To complement the learned classifier-based metrics, we directly measure physical motion intensity from the generated skeleton joint trajectories. 
        
        \textbf{Physical-Intensity Score.} First, we use root-relative range of motion (\textit{ROM}) as the primary physical-intensity measure, thereby reducing the influence of global character translation. Let \(\mathbf{x}_{i,t,j}\in\mathbb{R}^{3}\) denote the position of joint \(j\) at frame \(t\) in motion \(i\), and let \(r\) denote the root joint. We first compute the root-relative joint positions

        \vspace{-10pt}
        
        \begin{equation}
            \widetilde{\mathbf{x}}_{i,t,j}
            =
            \mathbf{x}_{i,t,j}
            -
            \mathbf{x}_{i,t,r}.
        \end{equation}

        % \vspace{-5pt}
    
        Then, the physical-intensity score for motion \(i\) is

        \vspace{-10pt}
    
        \begin{equation}
            s_i
            =
            \sum_{j=1}^{J}
            \left\|
                \operatorname*{max}_{t}
                \widetilde{\mathbf{x}}_{i,t,j}
                -
                \operatorname*{min}_{t}
                \widetilde{\mathbf{x}}_{i,t,j}
            \right\|_2,
        \end{equation}

        \vspace{-5pt}
    
        where the maximum and minimum are computed independently along each spatial coordinate. This score measures the accumulated spatial range traversed by the joints relative to the character root. Larger values indicate motions with greater pose and limb displacement.
    
        Let \(\mathcal{A}\) denote the set of evaluated actions, and let
        \(\mathcal{I}_a=\{i:a_i=a\}\) denote the set of samples corresponding to action \(a\). We measure whether increasing the requested exaggeration level produces a monotonic increase in physical intensity using

        \vspace{-10pt}
    
        \begin{equation}
            \rho_{\mathrm{phys}}
            =
            \frac{1}{|\mathcal{A}|}
            \sum_{a\in\mathcal{A}}
            \rho_{\mathrm{S}}\!\left(
                \{l_i\}_{i\in\mathcal{I}_a},
                \{s_i\}_{i\in\mathcal{I}_a}
            \right).
        \end{equation}

        \vspace{-5pt}
    
        The correlation is computed separately within each action and then averaged, preventing actions with naturally larger ranges of motion from dominating the result. A higher \(\rho_{\mathrm{phys}}\) indicates that increasing the requested exaggeration level more consistently produces greater physical motion intensity.
    
        \textbf{Physical-Intensity Separation.} Correlation measures the ordering of the intensity levels but does not indicate the magnitude of the change between them. We therefore additionally measure the standardized separation between the neutral and extremely exaggerated levels. For each action \(a\), we standardize its physical-intensity scores as

        \vspace{-10pt}
    
        \begin{equation}
            z_i
            =
            \frac{s_i-\mu_{a_i}}{\sigma_{a_i}},
        \end{equation}

        where \(\mu_a\) and \(\sigma_a\) are respectively the mean and standard deviation of the physical-intensity scores across all evaluated samples and exaggeration levels for action \(a\). Let

        \vspace{-5pt}
    
        \begin{equation}
            \bar z_{\ell}
            =
            \frac{1}{|\mathcal{I}_{\ell}|}
            \sum_{i\in\mathcal{I}_{\ell}} z_i,
            \qquad
            \mathcal{I}_{\ell}=\{i:l_i=\ell\},
        \end{equation}

        \vspace{-5pt}
        
        denote the mean standardized intensity at exaggeration level \(\ell\). We define the physical-intensity separation as

        \vspace{-20pt}
    
        \begin{equation}
            \Delta_{\mathrm{phys}}
            =
            \bar z_{4}-\bar z_{0}.
        \end{equation}

        % \vspace{-5pt}
    
        A larger \(\Delta_{\mathrm{phys}}\) indicates a greater physical separation between the neutral and extremely exaggerated motions. While \(\rho_{\mathrm{phys}}\) measures whether intensity increases consistently across the full exaggeration range, \(\Delta_{\mathrm{phys}}\) measures the magnitude of the difference between its two endpoints.
    
        \textbf{Reference-Based Motion Novelty.}
        Following the novelty principle introduced in AnimationBench~\citep{AnimationBench}, we measure how far an exaggerated motion deviates from its corresponding neutral reference in a learned motion-feature space. Let \(\mathbf{f}_{i}^{\mathrm{exag}}\) and \(\mathbf{f}_{i}^{\mathrm{neut}}\) denote the feature embeddings of an exaggerated motion and its paired neutral motion, respectively, extracted using the pretrained HumanML3D motion encoder. We compute

        \vspace{-5pt}
    
        \begin{equation}
            \mathrm{Nov}_i
            =
            1-
            \frac{
                \left\langle
                    \mathbf{f}_{i}^{\mathrm{exag}},
                    \mathbf{f}_{i}^{\mathrm{neut}}
                \right\rangle
            }{
                \left\|\mathbf{f}_{i}^{\mathrm{exag}}\right\|_2
                \left\|\mathbf{f}_{i}^{\mathrm{neut}}\right\|_2
            }.
        \end{equation}
    
        The paired motions use the same action condition and random seed, so that the score primarily reflects the change introduced by exaggeration rather than unrelated variation in generated content. We report the mean novelty of the highest requested exaggeration level relative to its corresponding neutral references. Larger values indicate a greater feature-space deviation from neutral motion. However, novelty measures the magnitude of this deviation and, by itself, does not determine whether the resulting motion is plausible or preserves the intended action.
    }

    \textbf{Summary.} The proposed metrics capture complementary aspects of controllable exaggeration. The classifier-based metrics, \(\mathrm{ELRA}@k\) and \(\rho_{\mathrm{clf}}\), measure whether generated motions are recognized as having the requested exaggeration levels. The classifier-free metrics, \(\rho_{\mathrm{phys}}\) and \(\Delta_{\mathrm{phys}}\), measure whether the requested progression is reflected in the ordering and magnitude of physical motion intensity. Finally, \(\mathrm{Nov.}\) measures the feature-space deviation of the exaggerated motion from its neutral reference. Together, these metrics evaluate exaggeration control independently of conventional motion-quality and action-preservation metrics, which we report separately.
}

\section{Remarks on Training-Free Exaggeration Approach}
\label{app:remarks_training_free_exaggeration}
{
    \vspace{-5pt}
    
    \textbf{Training-Free Exaggeration Formulation for Flow-Matching Models.} HY-Motion~\citep{HY-Motion} is a flow-matching motion model whose Diffusion Transformer (DiT) architecture \citep{DiffusionTransformer} predicts a velocity field that transports an initial Gaussian noise sample at flow time $t=0$ to a clean motion at $t=1$. Unlike MDM \citep{MDM} and KIMODO \citep{Kimodo}, which directly predict the clean motion during reverse diffusion, HY-Motion predicts the instantaneous direction in which the current noisy state should evolve. Therefore, applying our training-free guidance to HY-Motion requires first converting its velocity prediction into an estimate of the clean motion. 
        
    Given a text condition $c$ and initial noise $z_0\sim\mathcal{N}(0,I)$, we first use the frozen HY-Motion model to generate a neutral motion $Q_{\mathrm{neut}}$. Then, we apply the DMP-based exaggeration operator from Eq.~\ref{eq:dmp_exag_operator} to obtain the exaggerated target $Q_K=\mathcal{E}_{K}(Q_{\mathrm{neut}})$. Then, we repeat the flow-matching sampling process from the same initial noise $z_0$ used to generate $Q_{\mathrm{neut}}$. Let $z_t$ denote the current motion state at flow time $t$, and let $\mathcal{V}_{\theta}(z_t,t,c)$ denote the velocity predicted by the frozen HY-Motion model. Under the linear flow path

    \vspace{-20pt}
    
    \begin{equation}
        z_t=(1-t)z_0+tQ,
    \end{equation}
    the corresponding estimate of the clean motion at $t=1$ is
    
    \vspace{-10pt}
    
    \begin{equation}
        \label{eq:hymotion_clean_estimate}
        \widehat{Q}^{(t)}
        =
        z_t+(1-t)\mathcal{V}_{\theta}(z_t,t,c).
    \end{equation}
    
    \vspace{-5pt}
    
    This clean-motion estimate is the flow-matching analogue of the clean-motion prediction $\widehat{Q}^{(s)}$ used in the diffusion formulation (Sec.~\ref{sec:tf_formulation_diffusion}). Then, we guide this prediction toward the DMP-exaggerated target using
    
    \vspace{-12pt}
    
    \begin{equation}
        \label{eq:hymotion_tf_endpoint}
        \widetilde{Q}^{(t)}
        =
        \widehat{Q}^{(t)}
        +
        \omega_t\bigl(Q_K-\widehat{Q}^{(t)}\bigr),
        \qquad
        \omega_t=\omega\,\gamma(t),
    \end{equation}
    where $\omega$ is the overall guidance strength and $\gamma(t)\in[0,1]$ is the same ramp-and-taper schedule used in the diffusion formulation (Sec.~\ref{sec:tf_formulation_diffusion}). As with MDM and KIMODO, the neutral root motion is restored after the guidance update so that exaggeration changes the local body motion without altering the character's global trajectory. Because HY-Motion advances the sample using a velocity rather than a clean-motion prediction, the guided clean estimate is converted back into a velocity before the next integration step. For $t<1$, this gives
    \begin{equation}
        \label{eq:hymotion_tf_euler}
        \widetilde{V}^{(t)}
        =
        \frac{\widetilde{Q}^{(t)}-z_t}{1-t},
        \qquad
        z_{t+\Delta t}
        =
        z_t+\Delta t\,\widetilde{V}^{(t)}.
    \end{equation}
    Therefore, the DMP target modifies HY-Motion through its estimated clean endpoint. This means Eq.~\ref{eq:hymotion_tf_endpoint} moves the endpoint toward the desired exaggerated motion, and Eq.~\ref{eq:hymotion_tf_euler} translates that correction into the velocity used to advance the flow. In this way, subsequent evaluations of the frozen velocity model repeatedly reintroduce HY-Motion's learned motion prior. Consequently, the DMP provides the exaggeration direction while HY-Motion regularizes the motion throughout sampling, without requiring any update to the model parameters.

    \textbf{Exaggeration Control Knobs.} The training-free exaggeration method has two control knobs: $K$ determines the exaggeration level of $Q_K$ relative to $Q_{\mathrm{neut}}$, while $\omega$ controls how strongly the diffusion trajectory follows $Q_K$. Setting $K=1$ recovers the neutral DMP trajectory, while $\omega=0$ recovers the original sampling procedure. 
    
    \textbf{Character's Root (Global) Trajectory Lock.} Because our objective is mainly to exaggerate the local body motion rather than change where the character travels, the DMP does not amplify the character root's planar travel channels. However, during the guided diffusion process in the training-free approach, the model may alter character global movement to compensate for the increased local motion. To prevent such phenomena, we explicitly preserve the character root trajectory of the neutral sample by applying the original character’s root trajectory to the motion being guided during the training-free approach:

    \vspace{-10pt}
    
    \begin{equation}
        \widetilde{Q}^{(s)}_{\mathrm{root}}\leftarrow Q_{0,\mathrm{root}}
    \end{equation}
    
    after each guidance update in Eq.~\ref{eq:guide}. This keeps the global path consistent with the neutral motion while allowing the remaining body features to become progressively more expressive.

    \paragraph{Hyperparameters Default Values.} The training-free approach uses $15$ uniform keyframes and $N_w{=}50$ basis functions for the DMP-exaggerated motion ($Q_K$), and guidance weight $\omega{=}0.3$ for the iterative refinement process. Text classifier-free guidance (CFG) scale is $2.5$ for MDM, $5.0$ for HY-Motion, and $2.5$ for KIMODO.
}

\section{Remarks on Training-Time Exaggeration Approach}
\label{app:remarks_training-time_exaggeration}
{

    \subsection{Supervised Fine-Tuning Formulation for HY-Motion}
    \label{app:sft_formulation_hymotion}
    {
        HY-Motion~\citep{HY-Motion} is a flow-matching text-to-motion model. To formulate our training-time exaggeration approach (SFT) for this model, let $Q^{(\ell)}\in\mathbb{R}^{T\times D_{\mathrm{H}}}$ denote a motion at exaggeration level $\ell$ after conversion to HY-Motion's normalized native representation, where $D_{\mathrm{H}}=201$. Given $Q^{(\ell)}$, we sample Gaussian noise $\epsilon\sim\mathcal{N}(0,I)$ and a continuous flow time $t\sim\mathcal{U}[0,1]$, and construct the intermediate noisy state

        \vspace{-20pt}
        
        \begin{equation}
            \label{eq:hymotion_interpolation}
            z_t=(1-t)\epsilon+tQ^{(\ell)}.
        \end{equation}

        \vspace{-5pt}
        
        Under this linear flow path, the target (ground truth) velocity is
        
        \vspace{-10pt}
        
        \begin{equation}
            \label{eq:hymotion_velocity_target}
            V^{(\ell)}=Q^{(\ell)}-\epsilon.
        \end{equation}

        \vspace{-5pt}
    
        Let $c_{\ell}$ be the text condition describing the action and its exaggeration level. HY-Motion's diffusion transformer (DiT) \citep{DiffusionTransformer}, $\mathcal{V}_{\theta}$, predicts the velocity

        \vspace{-10pt}
        
        \begin{equation}
            \widehat{V}^{(\ell)}
            =
            \mathcal{V}_{\theta}(z_t,t,c_{\ell}).
        \end{equation}
        Then, we fine-tune the model using its native masked flow-matching objective:
        \begin{equation}
            \label{eq:sft_hymotion_loss}
            \mathcal{L}_{\mathrm{SFT}}^{\mathrm{HY}}
            =
            \mathbb{E}_{Q^{(\ell)},t,\epsilon}
            \left[
            \frac{1}{D_{\mathrm{H}}\sum_{i=1}^{T}M_i}
            \sum_{i=1}^{T}\sum_{d=1}^{D_{\mathrm{H}}}
            M_i
            \left(
                V^{(\ell)}-
                \mathcal{V}_{\theta}(z_t,t,c_{\ell})
            \right)_{i,d}^{2}
            \right],
        \end{equation}
        where \(M_i\in\{0,1\}\) is a temporal validity mask: \(M_i=1\) if frame \(i\) belongs to the original motion sequence and \(M_i=0\) if it was added as padding when batching variable-length motions. Consequently, padded frames do not contribute to the loss, and the denominator \(D_{\mathrm{H}}\sum_{i=1}^{T}M_i\) averages the squared error over only the valid frame--feature entries. 
        
        In this way, exaggeration is introduced through text conditioning $c_{\ell}$. We also apply classifier-free text dropout during fine-tuning so that classifier-free guidance remains available at inference time.
    }
    
    \subsection{Supervised Fine-Tuning Formulation for KIMODO}
    \label{app:sft_formulation_kimodo}
    {
        KIMODO~\citep{Kimodo} is a diffusion-based motion model with a two-stage denoiser and a clean-motion ($x_0$) prediction objective. Therefore, its supervised fine-tuning formulation follows the same forward-diffusion process as MDM in Eq.~\ref{eq:forward} (Sec.~\ref{sec:training_time_exaggeration}). Again, let $Q^{(\ell)}\in\mathbb{R}^{T\times D_{\mathrm{K}}}$ denote a motion at exaggeration level $\ell$ after conversion to KIMODO's normalized native representation, where $D_{\mathrm{K}}=273$. Therefore, at diffusion step $s$, the intermediate noisy motion latent is
        
        \begin{equation}
            \label{eq:kimodo_forward}
            z_s
            =
            \sqrt{\bar{\alpha}_s}\,Q^{(\ell)}
            +
            \sqrt{1-\bar{\alpha}_s}\,\epsilon,
            \qquad
            \epsilon\sim\mathcal{N}(0,I).
        \end{equation}
        
        Given the action-exaggeration text condition $c_{\ell}$, KIMODO predicts the clean motion directly:
        
        \begin{equation}
            \widehat{Q}^{(\ell)}
            =
            \mathcal{G}_{\theta}(z_s,s,c_{\ell}).
        \end{equation}

        \vspace{-5pt}
    
        The main difference from the MDM supervised fine-tuning formulation in Sec.~\ref{sec:training_time_exaggeration} is the reconstruction objective. KIMODO partitions its \(273\)-dimensional representation into six feature blocks $\mathcal{B}$:
        \begin{equation}
            \mathcal{B}
            =
            \left\{
            \begin{aligned}
                &b_{\mathrm{root}} &&(3),&
                &b_{\mathrm{heading}} &&(2),&
                &b_{\mathrm{pos}} &&(66),\\[-2pt]
                &b_{\mathrm{rot}} &&(132),&
                &b_{\mathrm{vel}} &&(66),&
                &b_{\mathrm{contact}} &&(4)
            \end{aligned}
            \right\}.
        \end{equation}
        
        These blocks respectively encode the smoothed root position $b_{\mathrm{root}}$, the global root heading in a two-dimensional cosine--sine representation $b_{\mathrm{heading}}$, the positions of \(22\) joints $b_{\mathrm{pos}}$, the global \(6\)D rotations of \(22\) joints $b_{\mathrm{rot}}$, their velocities $b_{\mathrm{vel}}$, and four foot-contact channels $b_{\mathrm{contact}}$. Their dimensions sum to \(3+2+66+132+66+4=273\).
        
        KIMODO applies an element-wise Smooth-\(L_1\) loss, also known as the Huber loss~\citep{Smooth_L1_Loss}, to each feature block. For the normalized scalar residual
        \(e_{i,d}=\widehat{Q}_{i,d}^{(\ell)}-Q_{i,d}^{(\ell)}\), the unit-threshold loss used in our implementation is
        \begin{equation}
            \operatorname{SmoothL1}(e_{i,d})
            =
            \begin{cases}
                \frac{1}{2}e_{i,d}^{2},
                & |e_{i,d}|<1,\\[2pt]
                |e_{i,d}|-\frac{1}{2},
                & |e_{i,d}|\geq 1.
            \end{cases}
        \end{equation}
        It is quadratic for small reconstruction errors and linear for large errors, making it less sensitive to large residuals than a purely squared-error objective.
        
        Let \(|b|\) denote the dimensionality of feature block \(b\). The masked blockwise reconstruction objective is
        \begin{equation}
            \label{eq:sft_kimodo_loss}
            \mathcal{L}_{\mathrm{SFT}}^{\mathrm{KIM}}
            =
            \mathbb{E}_{Q^{(\ell)},s,\epsilon}
            \left[
            \sum_{b\in\mathcal{B}}
            \frac{1}{|b|\sum_{i=1}^{T}M_i}
            \sum_{i=1}^{T}\sum_{d\in b}
            M_i\,
            \operatorname{SmoothL1}
            \left(
                \widehat{Q}_{i,d}^{(\ell)}
                -
                Q_{i,d}^{(\ell)}
            \right)
            \right].
        \end{equation}
        
        Here, \(M_i\) is the temporal validity mask defined above: it equals \(1\) for valid motion frames and \(0\) for padded frames. Dividing each block loss by \(|b|\) prevents higher-dimensional blocks, such as joint rotations, from dominating the objective solely because they contain more features. The block losses are equally weighted in our implementation.
    
        For both HY-Motion and KIMODO, we use the LoRA and prior-preservation strategy introduced in the main paper (Sec.~\ref{sec:training_time_exaggeration}). The pre-trained backbone weights remain frozen under LoRA, and the prior-preservation term applies the corresponding native objective (flow-matching velocity regression for HY-Motion and clean-motion reconstruction for KIMODO) to neutral motions generated by the frozen base model. This regularizes the learned exaggeration behavior toward the pre-trained motion prior.
    }

    \subsection{Motion Representations and Backbones}
    \label{app:motion_representations_backbones}
    {
        \vspace{-5pt}
        
        The three backbones (MDM, KIMODO, and HY-Motion) operate on per-frame motion features, but differ in their feature dimensions,
        skeleton conventions, frame rates, and generative objectives. Therefore, we fine-tune and guide each
        model directly in its native representation. For the shared text-motion evaluation, motions generated
        by HY-Motion and KIMODO are decoded to $22$-joint SMPL positions \citep{SMPL:2015} and re-encoded in the HumanML3D
        $263$-dimensional representation. Model-specific exaggeration control and physical metrics are computed from the corresponding native outputs. Table~\ref{table:motion_representations_backbones} summarizes the native representations used in our implementation.
        
        The captured motions in \datasetname~are first retargeted to a common SMPL-22 body skeleton \citep{SMPL:2015} and then encoded separately for each backbone. MDM uses the official HumanML3D representation
        \citep{Guo2022HumanML3D}, whereas HY-Motion and KIMODO use their native SMPL-H- and SMPL-X-based
        representations, respectively. In all cases, we normalize the resulting features using the statistics
        associated with the corresponding pre-trained checkpoint. This preserves the feature scale expected by
        each backbone during fine-tuning and inference.
        
        \vspace{-2pt}
        
        \textbf{MDM / HumanML3D-263.}
        MDM~\citep{MDM} is a transformer-encoder diffusion model with a frozen CLIP text encoder and a
        clean-motion ($x_0$) prediction objective. In our configuration, it uses a cosine noise schedule with
        $50$ diffusion steps and a text-conditioning dropout probability of $0.1$. Motion is represented at
        $20$~FPS using the $263$-dimensional HumanML3D \citep{Guo2022HumanML3D} feature vector containing root angular velocity ($1$), root linear velocity in the ground plane ($2$), root height ($1$), root-relative positions of the $21$ non-root
        joints ($63{=}21{\times}3$), continuous $6$D rotations of the $21$ non-root joints ($126{=}21{\times}6$)~\citep{Zhou2019}, local velocities of all $22$ joints ($66{=}22{\times}3$), and four foot-contact features.
        
        \vspace{-2pt}
        
        \textbf{HY-Motion / SMPL-H-201.}
        HY-Motion~\citep{HY-Motion} is a flow-matching DiT that predicts a velocity field
        $v_\theta(x_t,c,t)$. Given Gaussian noise $x_0$ and clean motion $x_1$, our SFT implementation uses the
        linear optimal-transport path $x_t=(1-t)x_0+t x_1$ and the target velocity $x_1-x_0$. HY-Motion
        represents motion at $30$~FPS using a $201$-dimensional feature vector containing root translation
        ($3$), root orientation in $6$D ($6$), local $6$D rotations for the $21$ non-root body joints
        ($126{=}21{\times}6$), and auxiliary positions for the $22$ body joints ($66{=}22{\times}3$). We use
        the released Lite ($0.46$B) and Standard ($1.0$B) variants. Text conditioning combines a global
        $768$-dimensional CLIP-L \citep{CLIP} embedding with $4096$-dimensional per-token Qwen3-8B \citep{Qwen3} features. At inference,
        we integrate the learned velocity field using $50$ Euler steps and a text classifier-free-guidance scale
        of $5.0$.
        
        \vspace{-2pt}
        
        \textbf{KIMODO / SMPL-X-273.} KIMODO~\citep{Kimodo} is a two-stage kinematic diffusion model: a root stage models a $5$-dimensional global-root block, and a body stage models the remaining $268$ dimensions. The model
        predicts clean motion ($x_0$) under a cosine DDPM schedule \citep{DDPM} with $T{=}1000$ training timesteps. Our inference uses $100$-step DDIM sampling \citep{DDIM}. We use the body-only SMPL-X variant, which represents $22$ joints at $30$~FPS with a $273$-dimensional feature vector containing a smoothed root
        position ($3$), global root heading represented by $[\cos,\sin]$ ($2$), joint-position features ($66{=}22{\times}3$), global $6$D joint rotations ($132{=}22{\times}6$), joint velocities ($66{=}22{\times}3$), and four foot-contact features. In the joint-position block, the horizontal coordinates are expressed relative to the smoothed root, while the vertical coordinate remains global. The checkpoint applies separate normalization statistics to the root block ($[0{:}5]$) and the body block ($[5{:}273]$). Text conditioning uses a $4096$-dimensional LLM2Vec embedding \citep{LLM2Vec} obtained from Meta-Llama-3-8B, and we use a text classifier-free-guidance scale of $2.5$ at inference.

        \vspace{-2pt}

        \textbf{Text Conditioning.} Each motion clip's caption combines an action phrase with an intensity phrase. The five intensity phrases are ``in a natural way'' ($\ell{=}0$), ``slightly exaggerated'' ($1$), ``moderately exaggerated'' ($2$), ``very exaggerated'' ($3$), and ``extremely exaggerated'' ($4$); e.g.\ ``a person kicks, very exaggerated''. To diversify supervision we build up to four captions per clip (one canonical phrasing plus up to three synonym combinations drawn from small action- and intensity-synonym lists) and sample one per training epoch. Therefore, exaggeration enters purely through the text prompt, so at inference the user communicates the exaggeration intensity by changing the phrase while keeping the action fixed.
        
        \begin{table}[t]
            \centering
            \caption{Native motion representations and generative parameterizations used by the three
            backbones including MDM, HY-Motion, and KIMODO. Motion feature dimensions are reported per frame.}
            \label{table:motion_representations_backbones}
            \small
            \resizebox{\linewidth}{!}{%
            \begin{tabular}{lccc}
                \toprule
                 & \textbf{MDM} & \textbf{HY-Motion} & \textbf{KIMODO} \\
                 & (HumanML3D) & (SMPL-H) & (SMPL-X) \\
                \midrule
                Feature dimension & $263$ & $201$ & $273$ \\
                Body joints & $22$ & $22$ & $22$ \\
                Frame rate & $20$~FPS & $30$~FPS & $30$~FPS \\
                Prediction target & clean motion ($x_0$) & velocity ($v$) & clean motion ($x_0$) \\
                Text encoder & CLIP & CLIP-L $+$ Qwen3-8B & LLM2Vec (Llama-3-8B) \\
                Training / sampling
                    & cosine DDPM, $50$ steps
                    & OT flow, $50$ Euler steps
                    & cosine DDPM ($T{=}1000$), $100$-step DDIM \\
                \midrule
                \multicolumn{4}{l}{\emph{Per-frame feature layout (dimensions)}} \\
                Root features
                    & angular vel. $1$, planar vel. $2$, height $1$
                    & translation $3$, orientation $6$
                    & position $3$, heading $2$ \\
                Joint-position features & $63$ & $66$ & $66$ \\
                Joint rotations ($6$D) & $126$ (local) & $126$ (local) & $132$ (global) \\
                Joint velocities & $66$ & --- & $66$ \\
                Foot contacts & $4$ & --- & $4$ \\
                \bottomrule
            \end{tabular}%
            }
        \end{table}

    }

    \subsection{Data Augmentation}
    \label{app:sft_data_augmentation}
    {
        \vspace{-5pt}
        
        Our graded-exaggeration dataset, \datasetname, is limited, so we expand it using three label-preserving augmentations. To this end, we first apply each augmentation to the common SMPL-$22$ joint trajectories \citep{SMPL:2015} before converting the resulting motion separately into the native representation of each motion model backbone. Specifically, we use the same backbone-specific preprocessing and feature-extraction pipeline used to construct its unaugmented training samples. This preserves the structural relationships among derived features, such as rotations, velocities, and foot-contact labels, which could become inconsistent if the native feature vectors were modified directly. In the following, we explain the three data augmentation techniques used to expand our \datasetname dataset.

        % \vspace{-10pt}
        
        \textbf{Left--Right Mirroring.}
        Following the standard augmentation used in HumanML3D~\citep{Guo2022HumanML3D}, we reflect each motion across the sagittal plane and exchange the corresponding left and right joints. This produces an anatomically consistent mirrored motion while preserving its action, exaggeration level, and text condition. We generate one mirrored variant of each motion window in the dataset.
        
        % \vspace{-10pt}
        
        \textbf{Position Jitter.}
        We generate an additional variant by adding small zero-mean Gaussian perturbations to the joint positions, using a standard deviation of \(1\)~cm. The perturbations are smoothed temporally to suppress high-frequency noise and preserve the overall action and exaggeration level.
        
        % \vspace{-10pt}
        
        \textbf{Balanced Windowing.}
        To prevent longer motion clips from contributing disproportionately more training samples, we draw approximately \(16\) distinct temporal windows from each motion clip. Window durations are sampled uniformly between \(2.0\) and \(9.8\) seconds, corresponding to \(40\)--\(196\) frames for MDM at \(20\)~FPS and \(60\)--\(294\) frames for HY-Motion and KIMODO at \(30\)~FPS. Motion clips shorter than the minimum window duration are kept as a single full-length window. Each selected window contributes its original, mirrored, and position-jittered variants.
    }

    \subsection{Implementation Details}
    \label{app:sft}
    {    
        We initialize each model from its public pre-trained checkpoint and retain its native architecture,
        motion representation, and generative objective. All losses are computed in the corresponding
        normalized feature space, and padded frames are excluded through temporal masks. Unless otherwise
        specified, we use the LoRA-based fine-tuning strategy described below and full fine-tuning is used only
        for the corresponding ablation in Tables~\ref{tab:mdm_sft_ablation}-\ref{tab:kimodo_sft_ablation}.
        
        \textbf{Low-Rank Adaptation.}
        For a pre-trained linear layer with weight matrix \(W\), LoRA~\citep{LoRA} introduces a trainable residual \(\Delta W=(\alpha/r)BA\), where \(r\) is the adapter rank. We initialize \(A\) using Kaiming-uniform initialization~\citep{kaiming_initialization}, which scales its random weights according to the input dimension, and initialize \(B\) to zero. Consequently, \(\Delta W=0\) at initialization, so the adapted model initially behaves identically to the pre-trained model. The pre-trained weights remain frozen, and only the adapter parameters are optimized. The adapted layers
        follow the structure of each backbone: MDM adapts the query, key, and value projections in the
        denoising transformer's attention layers; HY-Motion adapts the attention and MLP projections of its
        DiT; and KIMODO adapts the transformer feed-forward layers together with its motion input, motion
        output, and text projections. MDM sets $\alpha=r$. HY-Motion and KIMODO use $\alpha=16$, which gives
        unit scaling at their default rank $r=16$.

        \textbf{Prior Preservation.}
        To reduce forgetting of the pre-trained motion distribution, we augment the supervised exaggeration loss with a prior-preservation regularization term,
        \begin{equation}
            \mathcal{L}_{\mathrm{SFT}}
            =
            \mathcal{L}_{\mathrm{exag}}
            +
            \lambda_{\mathrm{prior}}\mathcal{L}_{\mathrm{prior}}.
        \end{equation}
        The prior term applies the same native training objective as the exaggeration term: clean-motion
        denoising for MDM (Eq.~\ref{eq:sft_diffusion_loss}) and KIMODO (Eq.~\ref{eq:sft_kimodo_loss}), and velocity regression for HY-Motion (Eq.~\ref{eq:sft_hymotion_loss}). For MDM, the reported
        experiments use real HumanML3D motions \cite{Guo2022HumanML3D} as prior examples. For HY-Motion and KIMODO, we generate a fixed set of neutral motions using the corresponding frozen base model and cache them before fine-tuning.
        
        \textbf{Optimization.}
        We optimize with AdamW \citep{AdamW} and apply classifier-free-conditioning dropout with probability $0.1$ using each backbone's native null condition. Gradients are clipped to a maximum norm of $1.0$. We maintain an exponential moving average of the trainable parameters for validation and checkpoint sampling,
        using decay $0.9999$ for MDM and $0.999$ for HY-Motion and KIMODO. Table~\ref{tab:app-sft} summarizes the default configurations used by our
        experiments.
        
        \begin{table}[t]
            \centering
            \caption{\textbf{Default supervised fine-tuning hyperparameters and configurations.}}
            \label{tab:app-sft}
            \small
            \resizebox{\linewidth}{!}{%
            \begin{tabular}{lccc}
                \toprule
                \textbf{Setting} & \textbf{MDM} & \textbf{HY-Motion} & \textbf{KIMODO} \\
                \midrule
                Objective & masked $x_0$ MSE & masked velocity MSE & masked $x_0$ smooth-$L_1$ \\
                Training steps & $16$k & $16$k & $16$k \\
                Batch size / GPU & $64$ & $16$ & $8$ \\
                Learning rate & $1\!\times\!10^{-4}$ & $1\!\times\!10^{-4}$ & $1\!\times\!10^{-4}$ \\
                Maximum frames & $196$ & $294$ & $300$ \\
                LoRA rank $r$ & $\{64,128\}$ & $\{16,64\}$ & $16$ \\
                LoRA scale $\alpha$ & $r$ & $16$ & $16$ \\
                CFG dropout & $0.1$ & $0.1$ & $0.1$ \\
                Gradient clipping & $1.0$ & $1.0$ & $1.0$ \\
                EMA decay & $0.9999$ & $0.999$ & $0.999$ \\
                $\lambda_{\mathrm{prior}}$ & $\{0.1,0.25\}$ & $\{0.1, 0.25\}$ & $0.25$ \\
                Prior source & HumanML3D & frozen-base samples & frozen-base samples \\
                Precision & FP32 & FP32 & FP32 \\
                \bottomrule
            \end{tabular}%
            }
        \end{table}
    }

}

\section{Additional Experiments and Results}
\label{app:additional_experiments_results}
{        
     We show expanded results, ablation studies, and discussions from the main paper featuring more controllable motion exaggeration examples across more diverse actions using our different approaches including TF, SFT, and SFT+TF in Figs.~\ref{fig:tradeoff}-~\ref{fig:result_controllable_exaggeration_tf} and Tables.~\ref{tab:mdm_sft_ablation}-~\ref{tab:ablation_exaggeration_components}.
         
    \textbf{Qualitative Results on More Diverse Actions.} We provide additional qualitative results to complement the representative comparisons in the main paper. First, Figs.~\ref{fig:result_sft+tf}--\ref{fig:result_tf} separately visualize motions generated using our SFT+TF, SFT, and TF approaches across a broader set of actions. These comparisons illustrate how each approach increases the expressiveness of the generated motion while retaining the structure of the underlying action. Second, Fig.~\ref{fig:result_mdm_hymotion} extends the qualitative evaluation to additional examples generated with the MDM and HY-Motion backbones, demonstrating that the proposed exaggeration framework is applicable across different motion representations and generative architectures. Finally, Figs.~\ref{fig:result_controllable_exaggeration_sft} and~\ref{fig:result_controllable_exaggeration_tf} illustrate the controllability of the proposed approaches. For SFT, exaggeration intensity is controlled through the intensity description included in the text condition, whereas for TF, it is controlled through the DMP exaggeration scale $K$. In both cases, increasing the corresponding control produces progressively more pronounced poses and motion amplitudes while preserving the identity of the original action.

    \textbf{Ablation studies on training-time (SFT) exaggeration approach.} The purpose of these ablations is to identify, for each motion-generation backbone, an SFT configuration that provides a balanced trade-off between exaggeration control and action preservation metrics. We evaluate exaggeration control using $\mathrm{ELRA}@1$, $\rho_{\mathrm{clf}}$, $\rho_{\mathrm{phys}}$, and $\Delta_{\mathrm{phys}}$ (explained in Appendix~\ref{app:remarks_exaggeration_evaluation_suit}), while action preservation and fidelity to the pre-trained motion distribution are assessed using R-precision, FID, Diversity, and MMDist following evaluation protocol used in \citep{Guo2022HumanML3D}. Additionally, Foot-Skate is reported to monitor motion quality. Across MDM, KIMODO, HY-Motion (0.4B), and HY-Motion (1B), we vary the fine-tuning strategy where applicable, the LoRA rank $r$, and the prior-preservation weight $\lambda$. For MDM, we additionally examine the source of the prior-preservation data. These factors control the effective adaptation strength. Specifically, greater adaptation capacity or weaker prior preservation can improve exaggeration control but may cause the generated motion to deviate from the original action and the pre-trained motion distribution, whereas stronger prior preservation can retain the original motion prior at the cost of suppressing exaggeration. Therefore, we do not select configurations by maximizing any single metric and instead we choose a balanced operating point that substantially improves exaggeration controllability while limiting degradation in action preservation and motion quality. Tables~\ref{tab:mdm_sft_ablation}-\ref{tab:kimodo_sft_ablation} report the complete results used for this selection.

    \textbf{User Study.} To perceptually assess the exaggeration capability of our method and its ability to preserve the underlying action, we conducted a user study with 20 participants. The controllable-exaggeration evaluation comprised 5 videos, each showing a neutral reference motion (level \(0\)) alongside three or four variants generated by our method at progressively increasing exaggeration levels. The examples used for this evaluation included walking, dancing, kicking, leaping forward, and jumping. The display order was permuted across examples, and the characters were identified only by letters so that the underlying exaggeration levels were not disclosed. For each video, participants selected both the most exaggerated motion and the most neutral motion, resulting in 10 questions per participant. We report the \emph{highest-intensity selection rate}, defined as the percentage of valid responses in which the largest requested exaggeration level was selected as the most exaggerated, and the \emph{neutral-reference selection rate}, defined as the percentage in which neutral reference motion was selected as the most neutral. For the \emph{action-preservation} evaluation, participants viewed a neutral source motion and its exaggerated counterpart and rated whether the counterpart preserved the type of action performed by the source character. The response options were \emph{very well preserved}, \emph{mostly preserved}, \emph{slightly preserved}, and \emph{not preserved}. We count the first two categories as positive responses. This evaluation comprised 6 questions. In total, each participant was presented with 16 questions. Fig.~\ref{fig:user_study_questionnaire} illustrates representative questions from the controllable-exaggeration and action-preservation evaluations presented to participants. As reported in Table~\ref{tab:user_study_full}, participants selected the highest requested exaggeration level as the most exaggerated in \(81.8\%\) of valid responses and selected the neutral reference motion as the most neutral in \(55\%\) of responses. The exaggerated motions received a positive action-preservation rating in \(76.7\%\) of cases. These results indicate that the highest requested exaggeration level was generally perceived as the most exaggerated while the underlying action was preserved in most evaluations.

    \begin{table*}[t]
        \centering
        \caption{\small \textbf{User-study results for controllable exaggeration and action preservation.} We report how often participants identified the highest requested intensity as the most exaggerated and the neutral reference motion as the most neutral. Action preservation is the percentage of ratings marked \emph{very well preserved} or \emph{mostly preserved}. Percentages are computed over valid responses, with counts shown in parentheses; ``--'' indicates that no example was evaluated.}
        \vspace{-5pt}
        \label{tab:user_study_full}
        \small
        \begin{tabular}{lccccc}
        \toprule
        Backbone & Exag. actions & Highest intensity & Neutral reference & Preserv. actions & Action preserved \\
        \midrule
        KIMODO & 3 & 86.7\% (52/60) & 80.0\% (48/60) & 5 & 82.0\% (82/100) \\
        HY-Motion & 1 & 78.9\% (15/19) & 10.0\% (2/20) & 1 & 50.0\% (10/20) \\
        MDM & 1 & 70.0\% (14/20) & 25.0\% (5/20) & 0 & -- \\
        \midrule
        \textbf{Overall} & 5 & 81.8\% (81/99) & 55.0\% (55/100) & 6 & 76.7\% (92/120) \\
        \bottomrule
        \end{tabular}
        \vspace{-5pt}
    \end{table*}
    
}

\section{The Use of Large Language Models}
{
    \vspace{-5pt}
    
    We used large language models (LLMs) for the following purposes: (i) editing, refining, and polishing the manuscript, including improving its grammar, wording, concision, organization, and overall readability; (ii) supporting retrieval and discovery during the literature review, particularly when searching for existing motion foundation models and prior work related to animation principles; and (iii) assisting with the presentation and verification of the work, including checking the consistency between manuscript descriptions and our implementation, and developing a script to aggregate the user-study responses and format the resulting statistics.
    
    LLMs were not used to generate the motion datasets or experimental results, run the reported experiments, or make final methodological or scientific decisions. All LLM-generated suggestions, including text, literature leads, and code, were carefully reviewed, verified, and, revised by the authors. Relevant literature was checked against the original sources, and all reported analyses and results were validated by the authors. The authors take full responsibility for the final manuscript and the accuracy of all its content, claims, results, and artifacts.
}

\clearpage

\begin{table*}[t]
  \centering

    \caption{\small \textbf{Ablation study of training-time (SFT) exaggeration approach on MDM.}
    We evaluate the effect of (a) fine-tuning strategy, (b) LoRA rank, (c) prior-preservation weight, and (d) prior-preservation data source across metrics measuring exaggeration control, prompt adherence, motion-manifold fidelity, and motion quality. The goal is to identify effective hyperparameters and configurations for supervised fine-tuning of the MDM model. Section (a) compares two fine-tuning strategies: (i) \emph{LoRA (ours)}, which trains LoRA adapters with the proposed prior-preservation objective, and (ii) \emph{Full FT}, which updates all model parameters using the same objective. Full fine-tuning produces stronger exaggeration, but substantially degrades prompt adherence and motion-manifold fidelity. In contrast, LoRA with prior preservation provides a better trade-off between exaggeration control and preservation of the base model's action semantics and motion manifold. In Sections (b) and (c), we ablate this LoRA+prior-preservation configuration by training identical models with different LoRA ranks $r$ and prior-preservation weights $\lambda$, respectively. Increasing the LoRA rank consistently improves exaggeration control but progressively reduces prompt adherence and manifold fidelity. Increasing the prior-preservation weight exhibits the opposite trend, improving prompt adherence and manifold fidelity while reducing the strength of exaggeration. Therefore, we select $r\in\{64,128\}$ and $\lambda\in\{0.1,0.25\}$, providing a practical trade-off between exaggeration and preservation. Finally, Section (d) compares different sources for the prior-preservation data: real HumanML3D motions, motions generated by the pre-trained MDM, and a mixture of the two. Using real HumanML3D motions generally provides stronger preservation of prompt adherence and motion-manifold fidelity, while generated or mixed priors provide comparable but slightly weaker results.}
    \label{tab:mdm_sft_ablation}
  % \small
  \resizebox{\textwidth}{!}{%
  \setlength{\tabcolsep}{5pt}
  \begin{tabular}{lccccccccccc}
    \toprule
    & \multicolumn{4}{c}{\textbf{Exaggeration}} & \multicolumn{6}{c}{\textbf{Action Preservation}} & \textbf{Quality} \\
    \cmidrule(lr){2-5} \cmidrule(lr){6-11} \cmidrule(lr){12-12}
    \textbf{Setting} & $\mathrm{ELRA}@1 \uparrow$ & $\rho_{\mathrm{clf}}\uparrow$ & $\rho_{\mathrm{phys}}\uparrow$ & $\Delta_{\mathrm{phys}}\uparrow$ & R@1$\uparrow$ & R@2$\uparrow$ & R@3$\uparrow$ & FID$\downarrow$ & Div$\rightarrow$ & MMDist$\downarrow$ & Skate$\downarrow$ \\
    \midrule
    \multicolumn{12}{l}{\emph{Reference rows}} \\
    \datasetname~ & 0.983 & 0.995 & 0.564 & 1.494 & -- & -- & -- & -- & -- & -- & 0.050 \\
    HumanML3D & -- & -- & -- & -- & 0.4530 & 0.6597 & 0.7675 & 0.0010 & 9.2461 & 3.2417 & 0.044 \\
    Base MDM (no FT) & 0.199 & 0.014 & 0.099 & 0.273 & 0.4019 & 0.6062 & 0.7211 & 0.4444 & 9.5169 & 3.5280 & 0.076 \\
    \midrule
    \multicolumn{12}{l}{\emph{(a) Fine-tuning strategy} 
    % (rank 5, $\lambda_{\mathrm{prior}}$=0.25, prior HumanML3D \textbf{vs.} full-model FT; $\mathrm{lr}=10^{-5}$)
    } \\
    LoRA (ours) & 0.227 & 0.165 & 0.393 & 1.139 & \textbf{0.3929} & \textbf{0.5882} & \textbf{0.6990} & \textbf{0.5652} & 9.4828 & \textbf{3.6276} & 0.086 \\
    Full FT & \textbf{0.342} & \textbf{0.530} & \textbf{0.744} & \textbf{2.160} & 0.2973 & 0.4520 & 0.5522 & 3.7578 & 8.9353 & 4.5992 & \textbf{0.082} \\
    \midrule
    \multicolumn{12}{l}{\emph{(b) LoRA rank} 
    % ($\lambda_{\mathrm{prior}}$=0.05, prior HumanML3D; $\mathrm{lr}=10^{-4}$)
    } \\
    $r$=1 & 0.203 & 0.036 & 0.120 & 0.318 & 0.4004 & \textbf{0.6025} & \textbf{0.7173} & 0.4747 & 9.4540 & \textbf{3.5344} & 0.087 \\
    $r$=4 & 0.202 & 0.009 & 0.144 & 0.365 & 0.4007 & 0.6000 & 0.7143 & 0.4616 & 9.3847 & 3.5560 & 0.094 \\
    $r$=8 & 0.208 & 0.043 & 0.265 & 0.732 & \textbf{0.4038} & 0.6002 & 0.7132 & \textbf{0.3833} & 9.3952 & 3.5489 & 0.095 \\
    $r$=16 & 0.219 & 0.097 & 0.369 & 1.075 & 0.4035 & 0.5975 & 0.7075 & 0.4441 & 9.4274 & 3.5787 & 0.090 \\
    $r$=32 & 0.232 & 0.182 & 0.421 & 1.212 & 0.3905 & 0.5828 & 0.6949 & 0.6024 & 9.3899 & 3.6576 & 0.084 \\
    $r$=64 & 0.256 & 0.276 & 0.442 & 1.257 & 0.3755 & 0.5602 & 0.6745 & 0.8684 & 9.3040 & 3.7732 & 0.078 \\
    $r$=128 & 0.301 & 0.370 & 0.511 & 1.522 & 0.3547 & 0.5405 & 0.6494 & 1.2379 & 9.2199 & 3.9219 & 0.072 \\
    $r$=256 & 0.326 & 0.423 & 0.586 & 1.738 & 0.3346 & 0.5168 & 0.6242 & 1.8154 & 9.1292 & 4.1080 & 0.069 \\
    $r$=512 & 0.339 & 0.433 & 0.646 & 1.879 & 0.3161 & 0.4855 & 0.5923 & 2.6873 & 8.9966 & 4.3296 & 0.066 \\
    $r$=768 & 0.340 & 0.447 & 0.676 & 1.995 & 0.3004 & 0.4653 & 0.5709 & 3.3625 & 8.9418 & 4.4975 & 0.065 \\
    $r$=1024 & \textbf{0.357} & \textbf{0.477} & \textbf{0.710} & \textbf{2.058} & 0.2916 & 0.4530 & 0.5590 & 3.4683 & 8.8735 & 4.5707 & \textbf{0.061} \\
    \midrule
    \multicolumn{12}{l}{\emph{(c) Prior-preservation weight} 
    % $\lambda_{\mathrm{prior}}$  (LoRA $r{=}1024$, prior HumanML3D; $\mathrm{lr}=10^{-4}$)
    } \\
    $\lambda$=0.01 & \textbf{0.367} & \textbf{0.500} & \textbf{0.734} & \textbf{2.150} & 0.2630 & 0.4108 & 0.5121 & 4.6972 & 8.7262 & 4.9157 & \textbf{0.059} \\
    $\lambda$=0.03 & 0.348 & 0.469 & 0.709 & 2.090 & 0.2816 & 0.4376 & 0.5411 & 4.0762 & 8.8212 & 4.7135 & 0.060 \\
    $\lambda$=0.05 & 0.357 & 0.477 & 0.710 & 2.058 & 0.2916 & 0.4530 & 0.5590 & 3.4683 & 8.8735 & 4.5707 & 0.061 \\
    $\lambda$=0.1 & 0.362 & 0.482 & 0.700 & 2.033 & 0.3042 & 0.4697 & 0.5752 & 3.2138 & 8.9436 & 4.4339 & 0.063 \\
    $\lambda$=0.25 & 0.346 & 0.439 & 0.687 & 2.000 & 0.3217 & 0.4958 & 0.6020 & 2.5860 & 8.9760 & 4.2150 & 0.065 \\
    $\lambda$=0.5 & 0.333 & 0.401 & 0.656 & 1.915 & 0.3445 & 0.5225 & 0.6333 & 1.9261 & 9.0553 & 4.0145 & 0.066 \\
    $\lambda$=0.75 & 0.324 & 0.362 & 0.626 & 1.842 & 0.3553 & 0.5419 & 0.6500 & 1.5578 & 9.1295 & 3.9030 & 0.067 \\
    $\lambda$=1 & 0.311 & 0.336 & 0.611 & 1.753 & 0.3652 & 0.5505 & 0.6616 & 1.3489 & 9.1905 & 3.8341 & 0.070 \\
    $\lambda$=5 & 0.240 & 0.098 & 0.471 & 1.401 & \textbf{0.3926} & \textbf{0.5875} & \textbf{0.7027} & \textbf{0.7383} & 9.7111 & \textbf{3.6403} & 0.091 \\
    \midrule
    \multicolumn{12}{l}{\emph{(d) Prior-preservation source} 
    % (LoRA, $\lambda_{\mathrm{prior}}$=0.25; $\mathrm{lr}=10^{-5}$)
    } \\
    HumanML3D & \textbf{0.217} & \textbf{0.142} & \textbf{0.410} & \textbf{1.211} & \textbf{0.3831} & \textbf{0.5736} & \textbf{0.6873} & 1.3971 & 9.0649 & \textbf{3.6922} & \textbf{0.086} \\
    Generated & 0.213 & 0.112 & 0.312 & 0.963 & 0.3690 & 0.5546 & 0.6671 & 1.6654 & 9.1062 & 3.8199 & 0.089 \\
    Mixed & 0.217 & 0.112 & 0.334 & 1.031 & 0.3804 & 0.5683 & 0.6824 & \textbf{1.3573} & 9.1361 & 3.7221 & 0.090 \\
    \bottomrule
  \end{tabular}
  }
\end{table*}

\begin{table*}[t]
  \centering
    \caption{\small \textbf{Ablation study of Training-time (SFT) exaggeration approach on HY-Motion (0.4B).}
    We evaluate the effect of (a) fine-tuning strategy, (b) LoRA rank, and (c) prior-preservation weight across metrics measuring exaggeration control, prompt adherence, motion-manifold fidelity, and motion quality. The goal is to identify effective hyperparameters and configurations for supervised fine-tuning of the HY-Motion (0.4B) model. Section (a) compares two fine-tuning strategies: (i) \emph{LoRA (ours)}, which trains LoRA adapters with the proposed prior-preservation objective, and (ii) \emph{Full FT}, which updates all model parameters without regularization. Full fine-tuning produces substantially stronger exaggeration but severely degrades prompt adherence and motion-manifold fidelity. In contrast, LoRA with prior preservation significantly improves exaggeration control while retaining the pretrained model's action semantics and motion manifold. In Section (b), we ablate the LoRA rank $r$ while keeping the prior-preservation weight fixed. Increasing $r$ improves exaggeration control from $r=1$ up to approximately $r=16$, after which the gains largely saturate. Meanwhile, prompt-adherence and manifold metrics remain relatively stable across ranks, with no consistent degradation as observed for MDM. We therefore select a relatively small rank, $r=16$, for HY-Motion (0.4B), which provides strong exaggeration control. In Section (c), we ablate the prior-preservation weight $\lambda$. Removing prior preservation ($\lambda=0$) yields the strongest classifier-based exaggeration, but substantially worsens prompt adherence and manifold fidelity. Increasing $\lambda$ generally improves prompt adherence, with the strongest R-precision obtained around $\lambda\in\{0.25,0.5\}$, while larger values eventually weaken exaggeration control. We therefore use $\lambda=0.25$ for the selected configuration, providing a strong balance between exaggeration and preservation of the pretrained motion prior.}
    \label{tab:hy0.4b_sft_ablation}
  \resizebox{\textwidth}{!}{%
  \setlength{\tabcolsep}{5pt}
  \begin{tabular}{lccccccccccc}
    \toprule
    & \multicolumn{4}{c}{\textbf{Exaggeration}} & \multicolumn{6}{c}{\textbf{Action Preservation}} & \textbf{Quality} \\
    \cmidrule(lr){2-5} \cmidrule(lr){6-11} \cmidrule(lr){12-12}
    \textbf{Setting} & $\mathrm{ELRA}@1 \uparrow$ & $\rho_{\mathrm{clf}}\uparrow$ & $\rho_{\mathrm{phys}}\uparrow$ & $\Delta_{\mathrm{phys}}\uparrow$ & R@1$\uparrow$ & R@2$\uparrow$ & R@3$\uparrow$ & FID$\downarrow$ & Div$\rightarrow$ & MMDist$\downarrow$ & Skate$\downarrow$ \\
    \midrule
    \multicolumn{12}{l}{\emph{Reference rows}} \\
    \datasetname~ & 0.995 & 0.999 & 0.542 & 1.534 & -- & -- & -- & -- & -- & -- & 0.071 \\
    HumanML3D & -- & -- & -- & -- & 0.4516 & 0.6596 & 0.7685 & 0.0010 & 9.0560 & 3.2341 & -- \\
    Base HY-Motion (no FT) & 0.335 & 0.453 & 0.335 & 0.970 & 0.3004 & 0.4617 & 0.5827 & 3.5259 & 8.1049 & 4.5667 & 0.046 \\
    \midrule
    \multicolumn{12}{l}{\emph{(a) Fine-tuning strategy} 
    % (LoRA + prior-preservation \textbf{vs.} unregularised full-model FT)
    } \\
    LoRA (ours) & 0.763 & 0.943 & 0.531 & 1.491 & \textbf{0.2883} & \textbf{0.4516} & \textbf{0.5706} & \textbf{4.7590} & 8.1070 & \textbf{4.6841} & 0.061 \\
    Full FT & \textbf{0.962} & \textbf{0.989} & \textbf{0.539} & \textbf{1.549} & 0.1724 & 0.2823 & 0.3679 & 12.6007 & 7.5104 & 6.1947 & \textbf{0.060} \\
    \midrule
    \multicolumn{12}{l}{\emph{(b) LoRA rank} 
    % ($\lambda_{\mathrm{prior}}$=0.1, prior HumanML3D)
    } \\
    $r$=1 & 0.653 & 0.889 & 0.552 & 1.568 & 0.3054 & 0.4657 & 0.5776 & \textbf{4.1510} & 7.8760 & 4.5865 & \textbf{0.054} \\
    $r$=4 & 0.700 & 0.915 & 0.559 & 1.562 & \textbf{0.3135} & \textbf{0.4778} & 0.5847 & 4.2825 & 7.9836 & \textbf{4.5031} & 0.061 \\
    $r$=8 & 0.738 & 0.931 & 0.527 & 1.502 & 0.3014 & 0.4748 & \textbf{0.5877} & 4.5399 & 7.8131 & 4.6309 & 0.059 \\
    $r$=16 & \textbf{0.768} & \textbf{0.942} & 0.501 & 1.444 & 0.2933 & 0.4758 & 0.5756 & 4.6824 & 8.1263 & 4.6606 & 0.058 \\
    $r$=32 & 0.748 & 0.938 & \textbf{0.561} & \textbf{1.586} & 0.2873 & 0.4698 & 0.5806 & 4.5664 & 7.9419 & 4.6014 & 0.057 \\
    $r$=64 & 0.765 & 0.933 & 0.523 & 1.485 & 0.2923 & 0.4667 & 0.5756 & 4.6874 & 8.1852 & 4.6542 & 0.061 \\
    $r$=128 & 0.752 & 0.934 & 0.523 & 1.466 & 0.2944 & 0.4516 & 0.5504 & 4.4536 & 8.1859 & 4.6502 & 0.058 \\
    $r$=256 & 0.752 & 0.931 & 0.530 & 1.529 & 0.2903 & 0.4587 & 0.5706 & 4.6146 & 7.9727 & 4.6222 & 0.059 \\
    $r$=512 & 0.703 & 0.931 & 0.539 & 1.525 & 0.2954 & 0.4667 & 0.5817 & 4.3785 & 8.4534 & 4.6506 & 0.059 \\
    \midrule
    \multicolumn{12}{l}{\emph{(c) Prior-preservation weight} 
    % $\lambda_{\mathrm{prior}}$ (LoRA=512, prior HumanML3D)
    } \\
    $\lambda$=0 & \textbf{0.839} & \textbf{0.955} & 0.527 & 1.494 & 0.2782 & 0.4284 & 0.5252 & 6.3173 & 8.1907 & 5.1383 & 0.059 \\
    $\lambda$=0.05 & 0.737 & 0.931 & 0.530 & 1.504 & 0.2933 & 0.4718 & 0.5867 & 4.4554 & 7.9255 & 4.5993 & 0.059 \\
    $\lambda$=0.1 & 0.736 & 0.933 & 0.527 & 1.475 & 0.2994 & 0.4526 & 0.5726 & 4.9073 & 8.1206 & 4.6936 & 0.058 \\
    $\lambda$=0.25 & 0.738 & 0.938 & 0.533 & 1.507 & 0.3135 & 0.4788 & 0.5817 & \textbf{4.3490} & 7.9348 & \textbf{4.5337} & 0.059 \\
    $\lambda$=0.5 & 0.724 & 0.923 & 0.526 & 1.485 & \textbf{0.3145} & 0.4788 & 0.5857 & 4.4406 & 8.0837 & 4.6461 & \textbf{0.054} \\
    $\lambda$=1 & 0.691 & 0.925 & \textbf{0.546} & \textbf{1.533} & 0.3125 & 0.4788 & 0.5857 & 4.5824 & 8.2268 & 4.6596 & 0.057 \\
    $\lambda$=5 & 0.527 & 0.833 & 0.502 & 1.407 & 0.3054 & \textbf{0.4909} & \textbf{0.6109} & 4.5394 & 8.3926 & 4.6541 & 0.058 \\
    \bottomrule
  \end{tabular}
  }
\end{table*}

\begin{table*}[t]
  \centering
    \caption{\small \textbf{Ablation study of training-time (SFT) exaggeration approach on HY-Motion (1B).}
    We evaluate the effect of (a) fine-tuning strategy, (b) LoRA rank, and (c) prior-preservation weight across metrics measuring exaggeration control, prompt adherence, motion-manifold fidelity, and motion quality. The goal is to identify effective hyperparameters and configurations for supervised fine-tuning of the HY-Motion (1B) model. Section (a) compares two fine-tuning strategies: (i) \emph{LoRA (ours)}, which trains LoRA adapters with the proposed prior-preservation objective, and (ii) \emph{Full FT}, which updates all model parameters without regularization. Full fine-tuning produces stronger exaggeration, but significantly degrades prompt adherence and motion-manifold fidelity. In contrast, LoRA with prior preservation maintains much stronger prompt adherence and manifold fidelity while still providing substantial exaggeration control. In Section (b), we ablate the LoRA rank $r$ with a fixed prior-preservation weight. Increasing $r$ generally improves classifier-based exaggeration control up to $r=128$. Prompt adherence and manifold fidelity also remain relatively stable across the tested ranks, with $r=256$ providing the strongest R-precision and MMDist among the rank ablations. We therefore select $r=64$ for the HY-Motion (1B) configuration, providing strong exaggeration control while retaining competitive prompt adherence and manifold fidelity without the additional capacity of larger adapters. In Section (c), we ablate the prior-preservation weight $\lambda$. Removing prior preservation ($\lambda=0$) gives the strongest classifier-based exaggeration, but significantly reduces prompt adherence and increases FID and MMDist. Introducing a small prior-preservation weight improves prompt adherence and manifold fidelity, with $\lambda=0.25$ providing particularly strong R-precision together with a high physical-intensity gap, while larger weights progressively weaken classifier-based exaggeration. We therefore use $\lambda=0.25$ for the selected configuration, providing a practical balance between exaggeration control and preservation of the pretrained motion prior.
    }
    \label{tab:hy1b_sft_ablation}
  % \small
  \resizebox{\textwidth}{!}{%
  \setlength{\tabcolsep}{5pt}
  \begin{tabular}{lccccccccccc}
    \toprule
    & \multicolumn{4}{c}{\textbf{Exaggeration}} & \multicolumn{6}{c}{\textbf{Action Preservation}} & \textbf{Quality} \\
    \cmidrule(lr){2-5} \cmidrule(lr){6-11} \cmidrule(lr){12-12}
    \textbf{Setting} & $\mathrm{ELRA}@1 \uparrow$ & $\rho_{\mathrm{clf}}\uparrow$ & $\rho_{\mathrm{phys}}\uparrow$ & $\Delta_{\mathrm{phys}}\uparrow$ & R@1$\uparrow$ & R@2$\uparrow$ & R@3$\uparrow$ & FID$\downarrow$ & Div$\rightarrow$ & MMDist$\downarrow$ & Skate$\downarrow$ \\
    \midrule
    \multicolumn{12}{l}{\emph{Reference rows}} \\
    \datasetname~ & 0.995 & 0.999 & 0.542 & 1.534 & -- & -- & -- & -- & -- & -- & 0.071 \\
    HumanML3D & -- & -- & -- & -- & 0.4527 & 0.6619 & 0.7719 & 0.0010 & 9.4287 & 3.2242 & -- \\
    Base HY-Motion 1B (no FT) & 0.291 & 0.346 & 0.300 & 0.720 & 0.3266 & 0.5262 & 0.6351 & 2.9705 & 8.2249 & 4.3312 & 0.044 \\
    \midrule
    \multicolumn{12}{l}{\emph{(a) Fine-tuning strategy} 
    % (LoRA + prior-preservation, $\lambda_{\mathrm{prior}}$=0.25 \textbf{vs.} unregularised full-model FT)
    } \\
    LoRA (ours) & 0.841 & 0.953 & \textbf{0.542} & 1.554 & \textbf{0.3276} & \textbf{0.5242} & \textbf{0.6310} & \textbf{3.5592} & \textbf{8.4329} & \textbf{4.3457} & 0.063 \\
    Full FT & \textbf{0.944} & \textbf{0.986} & 0.538 & \textbf{1.562} & 0.1734 & 0.3145 & 0.4032 & 5.8215 & 7.9433 & 5.5796 & \textbf{0.038} \\
    \midrule
    \multicolumn{12}{l}{\emph{(b) LoRA rank} 
    % ($\lambda_{\mathrm{prior}}$=0.1, prior HumanML3D)
    } \\
    $r$=1 & 0.730 & 0.920 & \textbf{0.609} & \textbf{1.740} & 0.3135 & 0.4909 & 0.5998 & 3.7862 & 7.8361 & 4.4610 & 0.062 \\
    $r$=4 & 0.792 & 0.946 & 0.559 & 1.570 & 0.3347 & 0.5192 & 0.6230 & 3.3686 & 8.5369 & 4.4019 & 0.062 \\
    $r$=8 & 0.818 & 0.955 & 0.547 & 1.543 & 0.3357 & 0.4889 & 0.6038 & 3.4149 & 8.2937 & 4.3819 & 0.060 \\
    $r$=16 & 0.813 & 0.941 & 0.535 & 1.523 & 0.3226 & 0.4879 & 0.6069 & \textbf{3.3473} & 8.0569 & 4.4192 & 0.063 \\
    $r$=32 & 0.835 & 0.948 & 0.542 & 1.548 & 0.3145 & 0.4909 & 0.6129 & 3.6444 & 8.1022 & 4.4219 & 0.062 \\
    $r$=64 & 0.850 & 0.957 & 0.521 & 1.509 & 0.3196 & 0.5121 & \textbf{0.6300} & 3.5303 & 8.1566 & 4.3980 & 0.062 \\
    $r$=128 & \textbf{0.855} & \textbf{0.963} & 0.521 & 1.484 & 0.3276 & 0.4970 & 0.5938 & 3.5654 & 8.2790 & 4.4507 & \textbf{0.058} \\
    $r$=256 & 0.844 & 0.954 & 0.540 & 1.544 & \textbf{0.3488} & \textbf{0.5212} & 0.6250 & 3.4028 & 8.4173 & \textbf{4.3071} & 0.064 \\
    \midrule
    \multicolumn{12}{l}{\emph{(c) Prior-preservation weight}
    % (LoRA=256, prior HumanML3D)
    } \\
    $\lambda$=0 & \textbf{0.889} & \textbf{0.972} & 0.536 & 1.526 & 0.2792 & 0.4315 & 0.5282 & 4.9065 & 8.2334 & 4.8583 & 0.065 \\
    $\lambda$=0.05 & 0.822 & 0.954 & 0.543 & 1.542 & 0.3145 & 0.4990 & 0.6099 & 3.4945 & 8.4698 & 4.4523 & 0.063 \\
    $\lambda$=0.1 & 0.833 & 0.949 & 0.540 & 1.536 & 0.3165 & 0.5020 & 0.6139 & \textbf{3.3388} & 8.0177 & 4.4065 & 0.062 \\
    $\lambda$=0.25 & 0.841 & 0.953 & 0.542 & \textbf{1.554} & 0.3276 & \textbf{0.5242} & \textbf{0.6310} & 3.5592 & 8.4329 & 4.3457 & 0.063 \\
    $\lambda$=0.5 & 0.806 & 0.945 & 0.531 & 1.522 & 0.3256 & 0.5212 & 0.6250 & 3.5808 & 8.4624 & 4.3903 & \textbf{0.059} \\
    $\lambda$=0.75 & 0.784 & 0.936 & \textbf{0.548} & 1.534 & 0.3135 & 0.5010 & 0.6290 & 3.5834 & 8.6047 & 4.3661 & 0.063 \\
    $\lambda$=1 & 0.803 & 0.942 & 0.545 & 1.532 & 0.3256 & 0.5091 & 0.6300 & 3.6054 & 8.4391 & 4.3395 & 0.063 \\
    $\lambda$=5 & 0.693 & 0.891 & 0.546 & 1.550 & \textbf{0.3387} & 0.5161 & 0.6260 & 3.4538 & 8.6014 & \textbf{4.3095} & 0.065 \\
    \bottomrule
  \end{tabular}
  }
\end{table*}

\begin{table*}[t]
  \centering
    \caption{\small \textbf{Ablation study of supervised fine-tuning (SFT) exaggeration approach on KIMODO.}
    We evaluate the effect of (a) LoRA rank and (b) prior-preservation weight across metrics measuring exaggeration control, prompt adherence, motion-manifold fidelity, and motion quality. The goal is to identify effective hyperparameters and configurations for supervised fine-tuning of the KIMODO model. In Section (a), we ablate the LoRA rank $r$ with a fixed prior-preservation weight. Unlike MDM and HY-Motion, in KIMODO, the effect of LoRA rank on exaggeration control is not monotonic: $\mathrm{ELRA}@1$ and $\rho_{\mathrm{clf}}$ improve from $r=1$ to moderate ranks but fluctuate thereafter, while the strongest physical-intensity correlation and neutral-to-extreme physical gap are obtained at the largest tested rank, $r=512$. Prompt adherence and manifold fidelity also vary non-monotonically across ranks, with different ranks providing the strongest individual metrics; for example, $r=256$ gives the highest R@1, while $r=32$ gives the lowest FID. These results suggest that increasing LoRA capacity does not uniformly improve either exaggeration control or prompt adherence for KIMODO. In Section (b), we ablate the prior-preservation weight $\lambda$. Removing prior preservation ($\lambda=0$) gives the strongest classifier-based exaggeration metrics, but substantially degrades prompt adherence and manifold fidelity. Introducing a small prior-preservation weight improves the text-motion alignment metrics, with $\lambda\in\{0.1,0.25\}$ providing strong R-precision and MMDist, while larger values generally reduce classifier-based exaggeration. The physical-intensity metrics remain relatively stable across moderate values of $\lambda$, with $\lambda=0.5$ achieving the largest $\Delta_{\mathrm{phys}}$. Based on these results, we select a moderate LoRA rank and a small prior-preservation weight to balance exaggeration control with preservation of the pretrained motion prior.
    }

    \label{tab:kimodo_sft_ablation}
  % \small
  \resizebox{\textwidth}{!}{%
  \setlength{\tabcolsep}{5pt}
  \begin{tabular}{lccccccccccc}
    \toprule
    & \multicolumn{4}{c}{\textbf{Exaggeration}} & \multicolumn{6}{c}{\textbf{Action Preservation}} & \textbf{Quality} \\
    \cmidrule(lr){2-5} \cmidrule(lr){6-11} \cmidrule(lr){12-12}
    \textbf{Setting} & $\mathrm{ELRA}@1\uparrow$ & $\rho_{\mathrm{clf}}\uparrow$ & $\rho_{\mathrm{phys}}\uparrow$ & $\Delta_{\mathrm{phys}}\uparrow$ & R@1$\uparrow$ & R@2$\uparrow$ & R@3$\uparrow$ & FID$\downarrow$ & Div$\rightarrow$ & MMDist$\downarrow$ & Skate$\downarrow$ \\
    \midrule
    \multicolumn{12}{l}{\emph{Reference rows}} \\
    \datasetname~ & 0.990 & 0.998 & 0.540 & 1.527 & -- & -- & -- & -- & -- & -- & 0.071 \\
    HumanML3D & -- & -- & -- & -- & 0.4537 & 0.6611 & 0.7748 & 0.0010 & 9.4376 & 3.2301 & -- \\
    Base KIMODO (no FT) & 0.203 & -0.024 & 0.174 & 0.448 & 0.3291 & 0.5373 & 0.6442 & 1.7365 & 8.9833 & 4.0029 & 0.026 \\
    \midrule
    \multicolumn{12}{l}{\emph{(a) LoRA rank} 
    % ($\lambda_{\mathrm{prior}}$=0.1, prior HumanML3D)
    } \\
    $r$=1 & 0.450 & 0.696 & 0.598 & 1.660 & 0.3377 & 0.5192 & 0.6290 & 1.3308 & 9.3574 & \textbf{4.0667} & \textbf{0.032} \\
    $r$=4 & 0.514 & 0.782 & 0.600 & 1.651 & 0.3387 & \textbf{0.5252} & 0.6240 & 1.2221 & 9.2667 & 4.0843 & 0.034 \\
    $r$=8 & 0.477 & 0.743 & 0.614 & 1.701 & 0.3306 & 0.5010 & 0.6038 & 1.3012 & 9.3138 & 4.1914 & 0.035 \\
    $r$=16 & \textbf{0.527} & 0.786 & 0.592 & 1.666 & 0.3306 & 0.5050 & \textbf{0.6331} & 1.2317 & 8.8285 & 4.1005 & 0.036 \\
    $r$=32 & 0.499 & 0.798 & 0.595 & 1.652 & 0.3306 & 0.5222 & 0.6159 & \textbf{1.0634} & 8.7674 & 4.0669 & 0.040 \\
    $r$=64 & 0.515 & \textbf{0.827} & 0.594 & 1.649 & 0.3427 & 0.4980 & 0.6280 & 1.1625 & 9.2137 & 4.1029 & 0.035 \\
    $r$=128 & 0.500 & 0.751 & 0.597 & 1.648 & 0.3337 & 0.5171 & 0.6139 & 1.2074 & 8.8401 & 4.0955 & 0.041 \\
    $r$=256 & 0.491 & 0.791 & 0.602 & 1.675 & \textbf{0.3438} & 0.5071 & 0.6139 & 1.2317 & 8.7661 & 4.1262 & 0.038 \\
    $r$=512 & 0.500 & 0.805 & \textbf{0.615} & \textbf{1.710} & 0.3347 & 0.4950 & 0.6149 & 1.2379 & 9.0761 & 4.1365 & 0.035 \\
    \midrule
    \multicolumn{12}{l}{\emph{(b) Prior-preservation weight} 
    % $\lambda_{\mathrm{prior}}$ (LoRA $r=64$, prior HumanML3D)
    } \\
    $\lambda$=0 & \textbf{0.525} & \textbf{0.819} & \textbf{0.604} & 1.655 & 0.3196 & 0.4677 & 0.5786 & 2.0470 & 9.0004 & 4.4328 & 0.042 \\
    $\lambda$=0.05 & 0.489 & 0.771 & 0.595 & 1.655 & 0.3256 & 0.5101 & 0.6179 & 1.1698 & 9.2136 & 4.1346 & 0.040 \\
    $\lambda$=0.1 & 0.492 & 0.744 & 0.600 & 1.653 & 0.3397 & \textbf{0.5242} & 0.6270 & 1.2117 & 8.9204 & 4.0824 & \textbf{0.037} \\
    $\lambda$=0.25 & 0.476 & 0.768 & 0.600 & 1.672 & \textbf{0.3438} & 0.5121 & 0.6280 & 1.1693 & 9.0286 & \textbf{4.0539} & 0.039 \\
    $\lambda$=0.5 & 0.498 & 0.744 & 0.599 & \textbf{1.691} & 0.3276 & 0.5040 & 0.6139 & \textbf{1.1671} & 8.9467 & 4.0736 & 0.040 \\
    $\lambda$=1 & 0.487 & 0.739 & \textbf{0.604} & 1.673 & 0.3226 & 0.5060 & \textbf{0.6341} & 1.3789 & 8.9422 & 4.0915 & 0.039 \\
    $\lambda$=5 & 0.412 & 0.687 & 0.569 & 1.615 & 0.3185 & 0.5091 & 0.6240 & 1.3913 & 9.0369 & 4.0880 & 0.042 \\
    \bottomrule
  \end{tabular}
  }
\end{table*}

    \begin{table*}[t]
    \vspace{-10pt}
        \centering
        \caption{\small \textbf{Contribution of the proposed motion-exaggeration approaches.} We compare the original HY-Motion-0.4B, HY-Motion-1B, MDM, and KIMODO baselines with DMP, training-free guidance (TF), supervised fine-tuning (SFT), and their composition (SFT+TF). Each approach is evaluated using its corresponding exaggeration control: text-conditioned intensity levels for the Baseline and SFT, increasing DMP scale factors for DMP and TF, and both controls jointly for SFT+TF. The metrics evaluate exaggeration control, prompt adherence and motion-manifold fidelity, and motion quality, revealing the trade-off between stronger exaggeration and preservation of the underlying action. ``\datasetname'' provides a reference for the exaggeration-control metrics, while ``HumanML3D'' provides a reference for the standard text-to-motion metrics; neither reference row is included when identifying the best method. \(\uparrow\) and \(\downarrow\) indicate that higher and lower values are better, respectively. The best result among the generated-motion methods for each backbone and metric is shown in bold.}
        \label{tab:ablation_exaggeration_components}
        \resizebox{\textwidth}{!}{%
        \begin{tabular}{llcccccccccccc}
        \toprule
        
        & & \multicolumn{5}{c}{\textbf{Exaggeration control}}
        & \multicolumn{6}{c}{\textbf{Prompt adherence \& manifold}}
        & \multicolumn{1}{c}{\textbf{Quality}} \\
        
        \cmidrule(lr){3-7}
        \cmidrule(lr){8-13}
        \cmidrule(lr){14-14}
        
        \textbf{Model} & \textbf{Method}
        & $\mathrm{ELRA}@1\uparrow$
        & $\rho_{\mathrm{clf}}\uparrow$
        & $\rho_{\mathrm{phys}}\uparrow$
        & $\Delta_{\mathrm{phys}}\uparrow$
        & Nov.$\uparrow$
        & R@1$\uparrow$
        & R@2$\uparrow$
        & R@3$\uparrow$
        & FID$\downarrow$
        & Div$\uparrow$
        & MMDist$\downarrow$
        & Skate$\downarrow$ \\
        
        \midrule
        
        \multirow{1}{*}{\textbf{HY-Motion-0.4B}}    
            & \datasetname
            & 0.995 & 0.999 & 0.542 & 1.534 & --
            & -- & -- & -- & -- & -- & -- & -- \\
        
            & HumanML3D
            & -- & -- & -- & -- & --
            & 0.4481 & 0.6554 & 0.7640 & 0.0011 & 9.3215 & 3.2445 & -- \\
        
            & Baseline
            & 0.331 & 0.267 & 0.315 & 0.961 & 0.069
            & \textbf{0.3945} & \textbf{0.6016} & \textbf{0.7305}
            & \textbf{2.5235} & 8.3203 & \textbf{3.5885} & 0.051 \\
        
            & \hspace{0.5em} + DMP
            & 0.231 & 0.055 & 0.471 & 1.257 & 0.185
            & 0.1875 & 0.2969 & 0.3828 & 34.5822 & 4.6140 & 6.2699 & 0.080 \\
        
            & \hspace{0.5em} + TF
            & 0.223 & 0.039 & 0.366 & 0.947 & 0.107
            & 0.2383 & 0.3867 & 0.5273 & 22.0843 & 5.9314 & 5.1676 & 0.072 \\
        
            & \hspace{0.5em} + SFT
            & \textbf{0.923} & \textbf{0.982} & 0.554 & 1.573 & 0.214
            & 0.3320 & 0.5625 & 0.6836 & 3.0531 & 8.2520 & 3.9408 & \textbf{0.042} \\
        
            & \hspace{0.5em} + SFT+TF
            & 0.892 & 0.975 & \textbf{0.708} & \textbf{1.997} & \textbf{0.309}
            & 0.2227 & 0.3711 & 0.5039 & 21.1181 & 5.7263 & 5.2920 & 0.067 \\
        
        \midrule
        
        \multirow{1}{*}{\textbf{HY-Motion-1B}}
            & \datasetname
            & 0.995 & 0.999 & 0.542 & 1.534 & --
            & -- & -- & -- & -- & -- & -- & -- \\
        
            & HumanML3D
            & -- & -- & -- & -- & --
            & 0.4582 & 0.6631 & 0.7681 & 0.0010 & 9.4410 & 3.2359 & -- \\
        
            & Baseline
            & 0.354 & 0.354 & 0.391 & 0.902 & 0.135
            & \textbf{0.4414} & \textbf{0.6484} & \textbf{0.7422}
            & \textbf{2.0825} & 8.9325 & \textbf{3.4851} & \textbf{0.029} \\
        
            & \hspace{0.5em} + DMP
            & 0.200 & -0.042 & 0.516 & 1.318 & 0.167
            & 0.1523 & 0.2969 & 0.4023 & 30.8515 & 4.7568 & 6.0145 & 0.061 \\
        
            & \hspace{0.5em} + TF
            & 0.200 & 0.007 & 0.424 & 1.017 & 0.082
            & 0.2266 & 0.4336 & 0.5898 & 17.0509 & 6.4124 & 4.8573 & 0.057 \\
        
            & \hspace{0.5em} + SFT
            & \textbf{0.938} & \textbf{0.985} & 0.556 & 1.622 & 0.222
            & 0.4102 & 0.6367 & 0.7383 & 2.2297 & 8.4439 & 3.6413 & 0.040 \\
        
            & \hspace{0.5em} + SFT+TF
            & 0.900 & 0.973 & \textbf{0.697} & \textbf{2.003} & \textbf{0.301}
            & 0.2812 & 0.4297 & 0.5430 & 19.7208 & 5.8073 & 5.1643 & 0.063 \\
        
        \midrule
        
        \multirow{1}{*}{\textbf{MDM}}    
            & \datasetname
            & 0.983 & 0.995 & 0.564 & 1.494 & --
            & -- & -- & -- & -- & -- & -- & 0.050 \\
        
            & HumanML3D
            & -- & -- & -- & -- & --
            & 0.4532 & 0.6558 & 0.7690 & 0.0011 & 9.2666 & 3.2277 & 0.044 \\

            & Baseline
            & 0.200 & 0.000 & 0.026 & 0.175 & 0.117
            & \textbf{0.4102} & \textbf{0.5898} & 0.7031
            & \textbf{0.8388} & 9.5722 & \textbf{3.6133} & \textbf{0.070} \\
        
            & \hspace{0.5em} + DMP
            & 0.200 & 0.000 & 0.358 & 0.953 & 0.130
            & 0.2734 & 0.4219 & 0.5625 & 20.9116 & 6.8088 & 5.2530 & 0.134 \\
        
            & \hspace{0.5em} + TF
            & 0.200 & 0.000 & 0.252 & 0.552 & 0.023
            & 0.3594 & 0.5859 & \textbf{0.7070} & 5.7547 & 8.3935 & 3.9870 & 0.091 \\
        
            & \hspace{0.5em} + SFT
            & \textbf{0.408} & 0.635 & 0.399 & 1.173 & 0.168
            & 0.2969 & 0.4844 & 0.5547 & 1.6349 & 8.8127 & 4.6498 & 0.073 \\
        
            & \hspace{0.5em} + SFT+TF
            & \textbf{0.408} & \textbf{0.643} & \textbf{0.545} & \textbf{1.610} & \textbf{0.276}
            & 0.2383 & 0.3789 & 0.5273 & 5.8575 & 7.7830 & 5.0234 & 0.085 \\
        
        \midrule
        
        \multirow{1}{*}{\textbf{KIMODO}}
            & \datasetname
            & 0.990 & 0.998 & 0.540 & 1.527 & --
            & -- & -- & -- & -- & -- & -- & 0.071 \\
        
            & HumanML3D
            & -- & -- & -- & -- & --
            & 0.4591 & 0.6649 & 0.7744 & 0.0010 & 9.3338 & 3.2309 & -- \\
        
            & Baseline
            & 0.223 & -0.051 & 0.137 & 0.281 & 0.080
            & \textbf{0.3555} & \textbf{0.5547} & \textbf{0.6562}
            & 2.0376 & 8.5198 & \textbf{3.8190} & \textbf{0.025} \\
        
            & \hspace{0.5em} + DMP
            & 0.208 & 0.053 & 0.374 & 1.086 & 0.144
            & 0.1445 & 0.2773 & 0.3516 & 38.1156 & 5.7478 & 6.7370 & 0.069 \\

            & \hspace{0.5em} + TF
            & 0.208 & 0.039 & 0.287 & 0.801 & 0.081
            & 0.2148 & 0.3516 & 0.4961 & 20.0688 & 6.8045 & 5.3506 & 0.036 \\
        
            & \hspace{0.5em} + SFT
            & 0.515 & 0.814 & 0.587 & 1.610 & 0.183
            & 0.3320 & 0.5234 & 0.6289 & \textbf{1.6934} & 8.9660 & 3.9654 & 0.027 \\
        
            & \hspace{0.5em} + SFT+TF
            & \textbf{0.531} & \textbf{0.831} & \textbf{0.729} & \textbf{2.015} & \textbf{0.306}
            & 0.1992 & 0.3594 & 0.4648 & 18.9754 & 7.2581 & 5.5249 & 0.030 \\
        
        \bottomrule
        \end{tabular}}
    \end{table*}

    \begin{figure}[t]
        \centering
        % --- (a) classifier-based exaggeration -------------------------------
        \begin{subfigure}{\textwidth}
            \centering
            \resizebox{0.98\textwidth}{!}{% <- change \textwidth (e.g. 0.9\textwidth) to shrink/grow
                \includegraphics{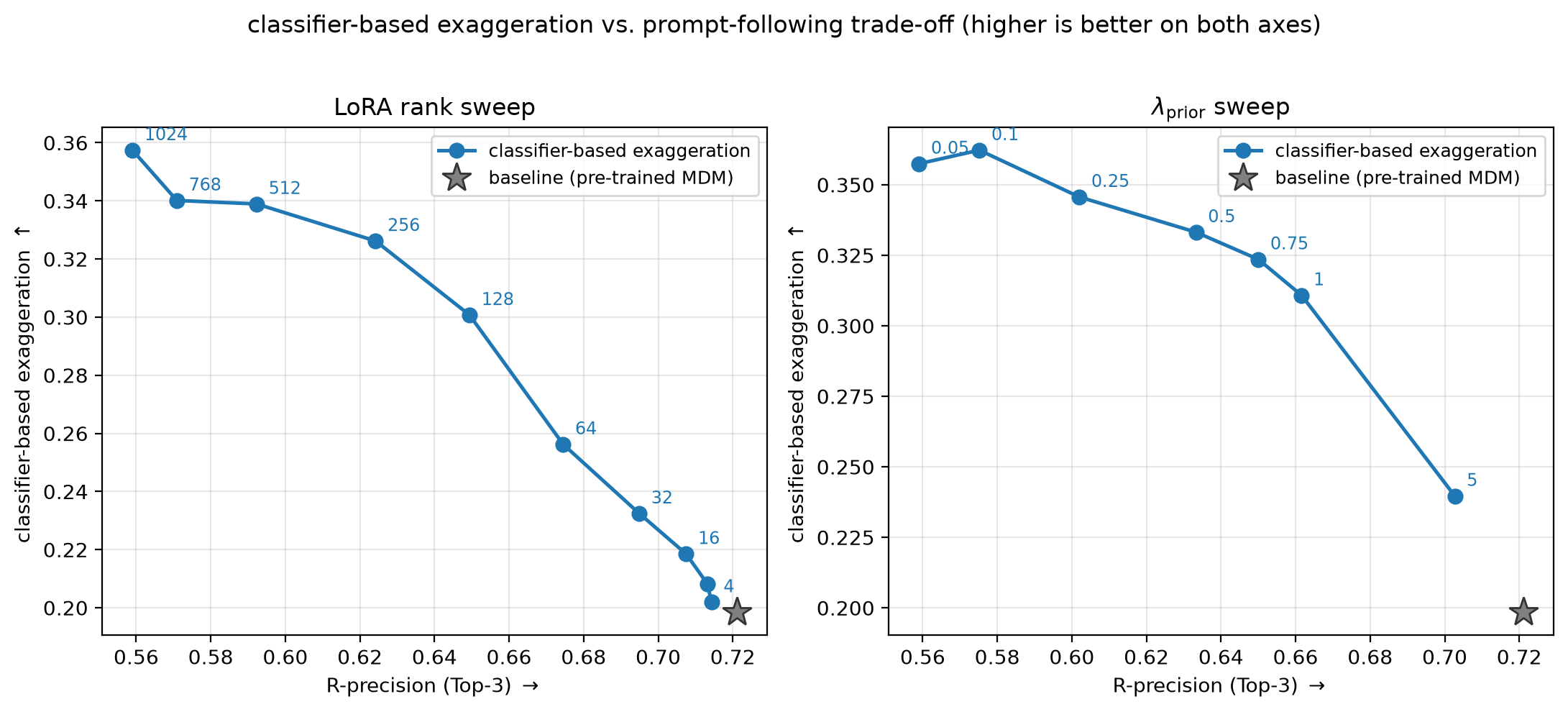}%
            }
            \caption{\small Ablation on $r$ and $\lambda$ using Classifier-based exaggeration and R-Precision metrics.}
            \label{fig:tradeoff-clf-based}
        \end{subfigure}
    
        \vspace{1.5em}
    
        % --- (b) classifier-free exaggeration --------------------------------
        \begin{subfigure}{\textwidth}
            \centering
            \resizebox{0.98\textwidth}{!}{%
                \includegraphics{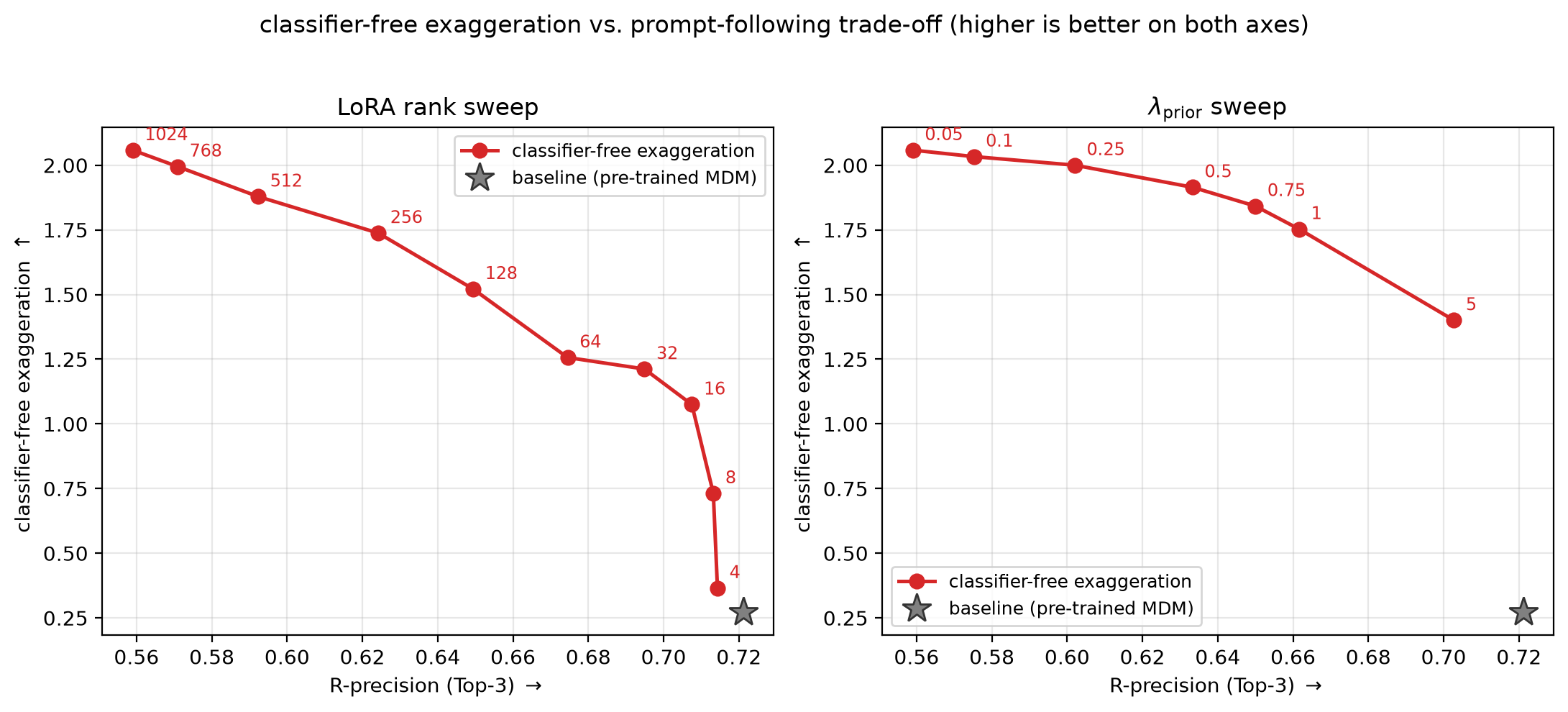}%
            }
            \caption{\small Ablation on $r$ and $\lambda$ using Classifier-free exaggeration and R-Precision metrics.}
            \label{fig:tradeoff-clf-free}
        \end{subfigure}
    
            \caption{\textbf{Ablation study of LoRA rank and prior-preservation weight for MDM.} We evaluate the trade-off between exaggeration control and prompt adherence by varying the LoRA rank $r$ (left) and prior-preservation weight $\lambda$ (right) for MDM \citep{MDM}. Higher $\mathrm{ELRA}@1$ and $\Delta_\mathrm{phys}$ indicate stronger exaggeration control, while higher R-precision indicates better prompt adherence. The pre-trained MDM baseline is denoted by a star.}
        \label{fig:tradeoff}
    \end{figure}

    \begin{figure}[t]
    \vspace{-4pt}
        \centering
        \includegraphics[width=0.98\linewidth]{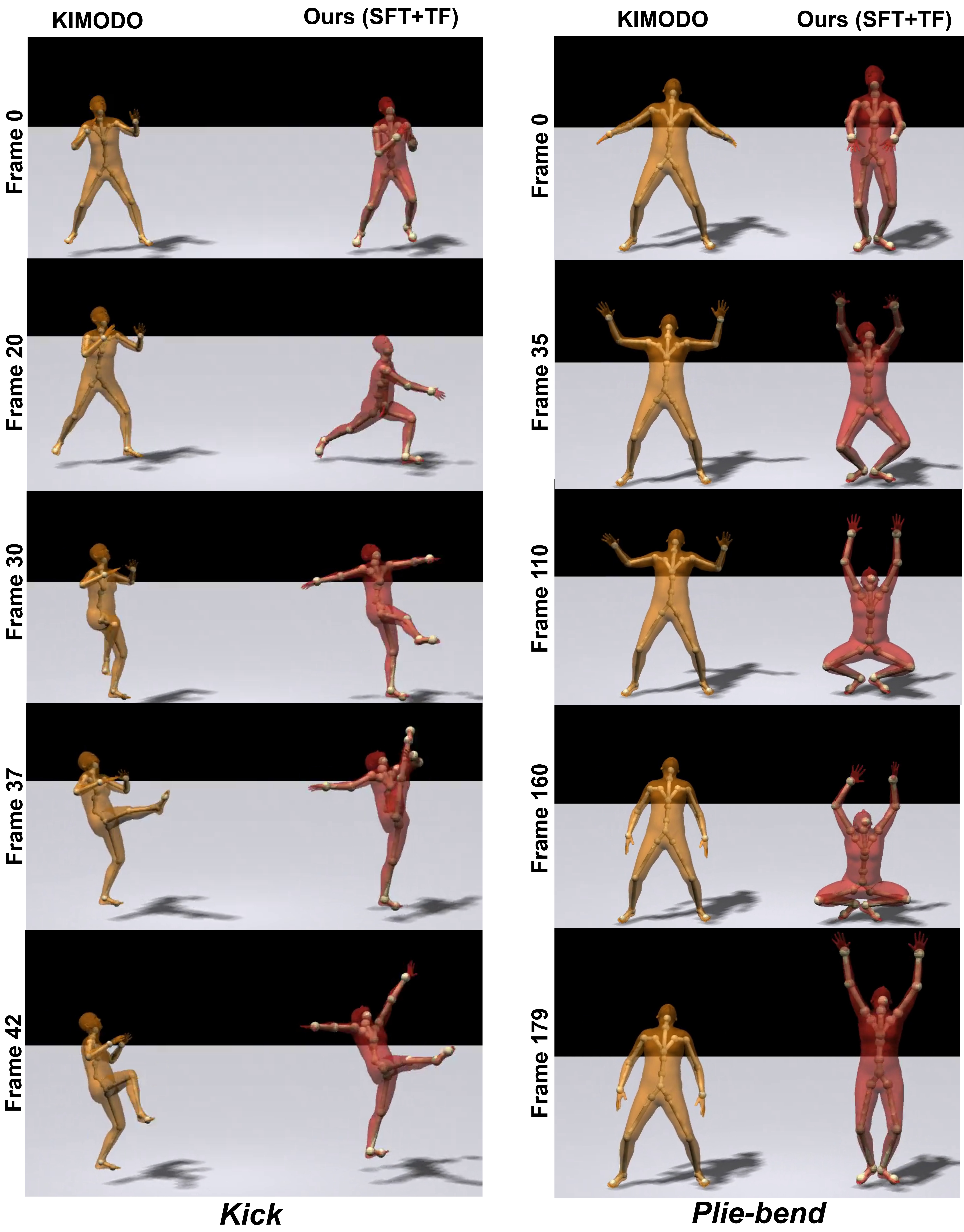}
        \vspace{-2pt}
        \caption{\textbf{Qualitative evaluation of our SFT+TF approach compared to KIMODO.} Selected frames from \textit{Kick} and \textit{Plie-bend} actions show that SFT+TF amplifies the characteristic poses and motion amplitudes of the original actions generated by KIMODO, producing more expressive exaggeration while preserving the underlying neutral action and its temporal progression.}
        \vspace{-13pt}
        \label{fig:result_sft+tf}
    \end{figure}

    \begin{figure}[t]
    \vspace{-20pt}
        \centering
        \includegraphics[width=0.98\linewidth]{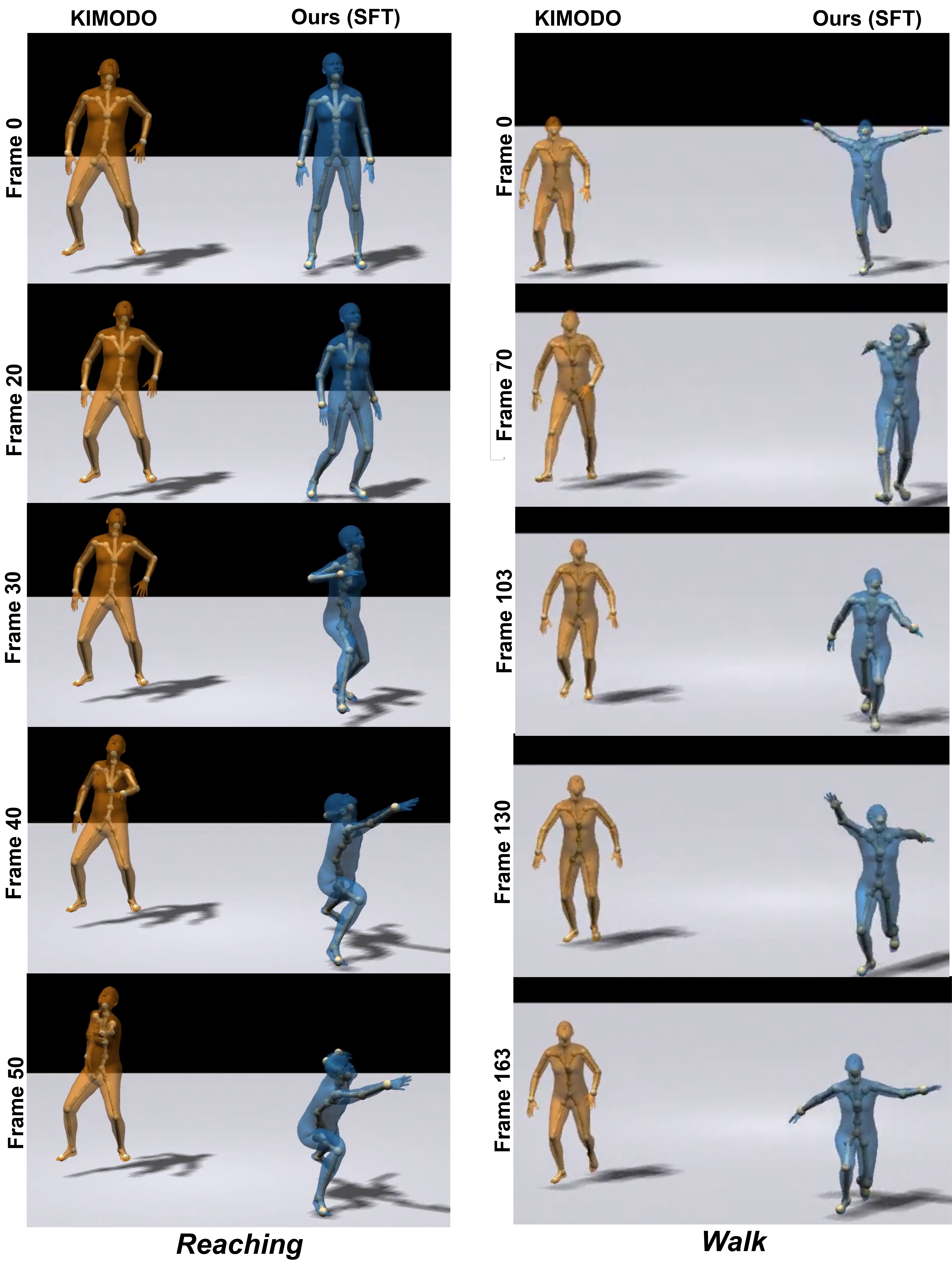}
        % \vspace{-2pt}
        \caption{\textbf{Qualitative evaluation of our SFT approach compared to KIMODO.} Selected frames from \textit{Reaching} and \textit{Walk} actions show that SFT amplifies the characteristic poses and motion amplitudes of the original actions generated by KIMODO, producing more expressive exaggeration while preserving the underlying neutral action and its temporal progression.}
        \vspace{-13pt}
        \label{fig:result_sft}
    \end{figure}

    \begin{figure}[t]
    \vspace{-35pt}
        \centering
        \includegraphics[width=0.98\linewidth]{Images/KMIODO_TF_Qualitative_Comparisons.jpg}
        % \vspace{-2pt}
        \caption{\textbf{Qualitative evaluation of our TF approach compared to KIMODO.} Selected frames from \textit{Dance} and \textit{Throw} actions show that TF amplifies the characteristic poses and motion amplitudes of the original actions generated by KIMODO, producing more expressive exaggeration while preserving the underlying neutral action and its temporal progression.}
        \vspace{-13pt}
        \label{fig:result_tf}
    \end{figure}

    \begin{figure}[t]
    \vspace{-4pt}
        \centering
        \includegraphics[width=0.98\linewidth]{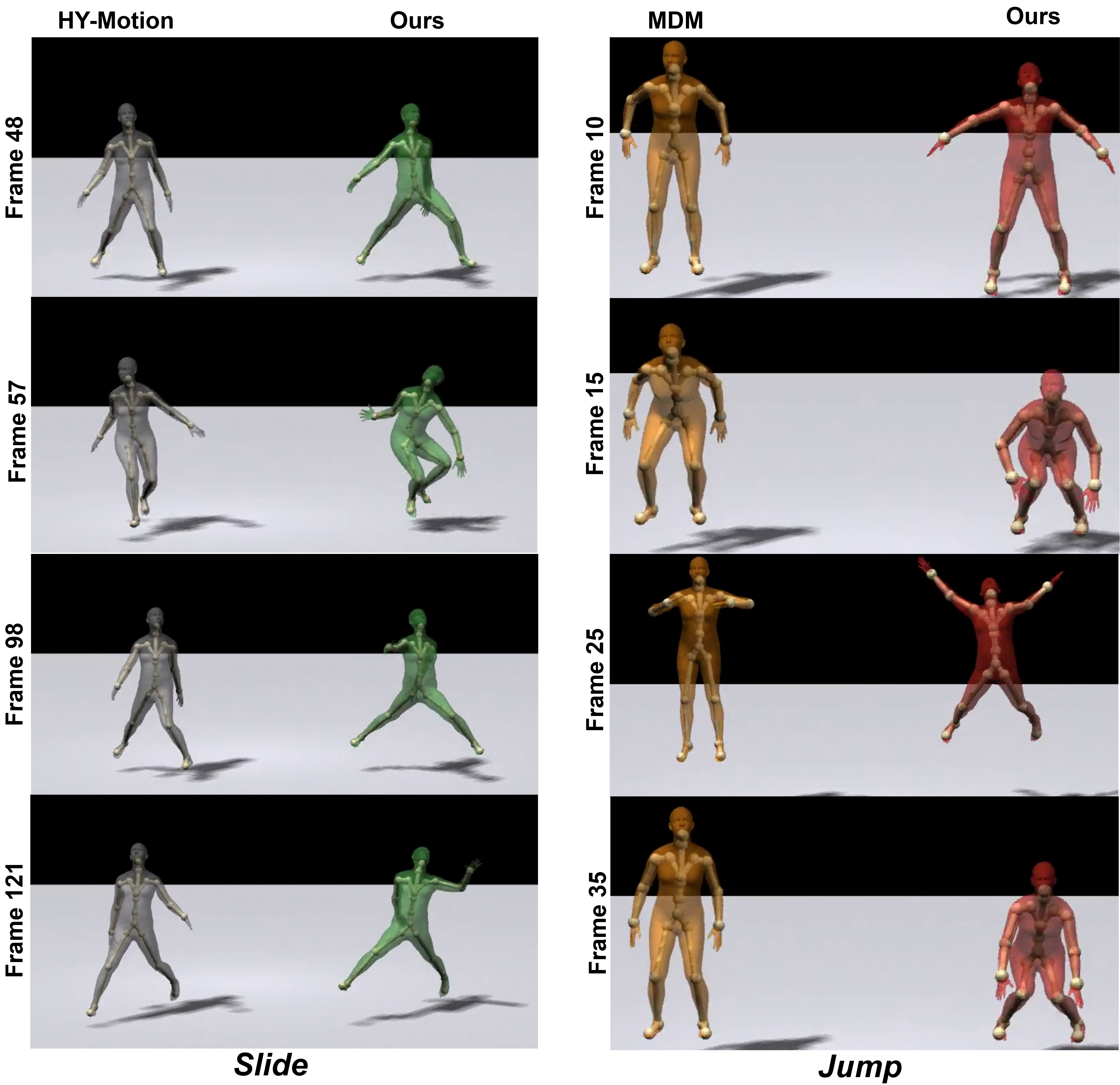}
        % \vspace{-2pt}
        \caption{\textbf{More qualitative results on MDM and HY-Motion.} Selected frames from \textit{Slide} and \textit{Jump} actions compare the baseline models with our approach. Across both motion generation models, our method amplifies characteristic poses and motion amplitudes while preserving the underlying neutral action and temporal progression.}
        \vspace{-13pt}
        \label{fig:result_mdm_hymotion}
    \end{figure}

    \begin{figure}[t]
    \vspace{-4pt}
        \centering
        \includegraphics[width=0.98\linewidth]{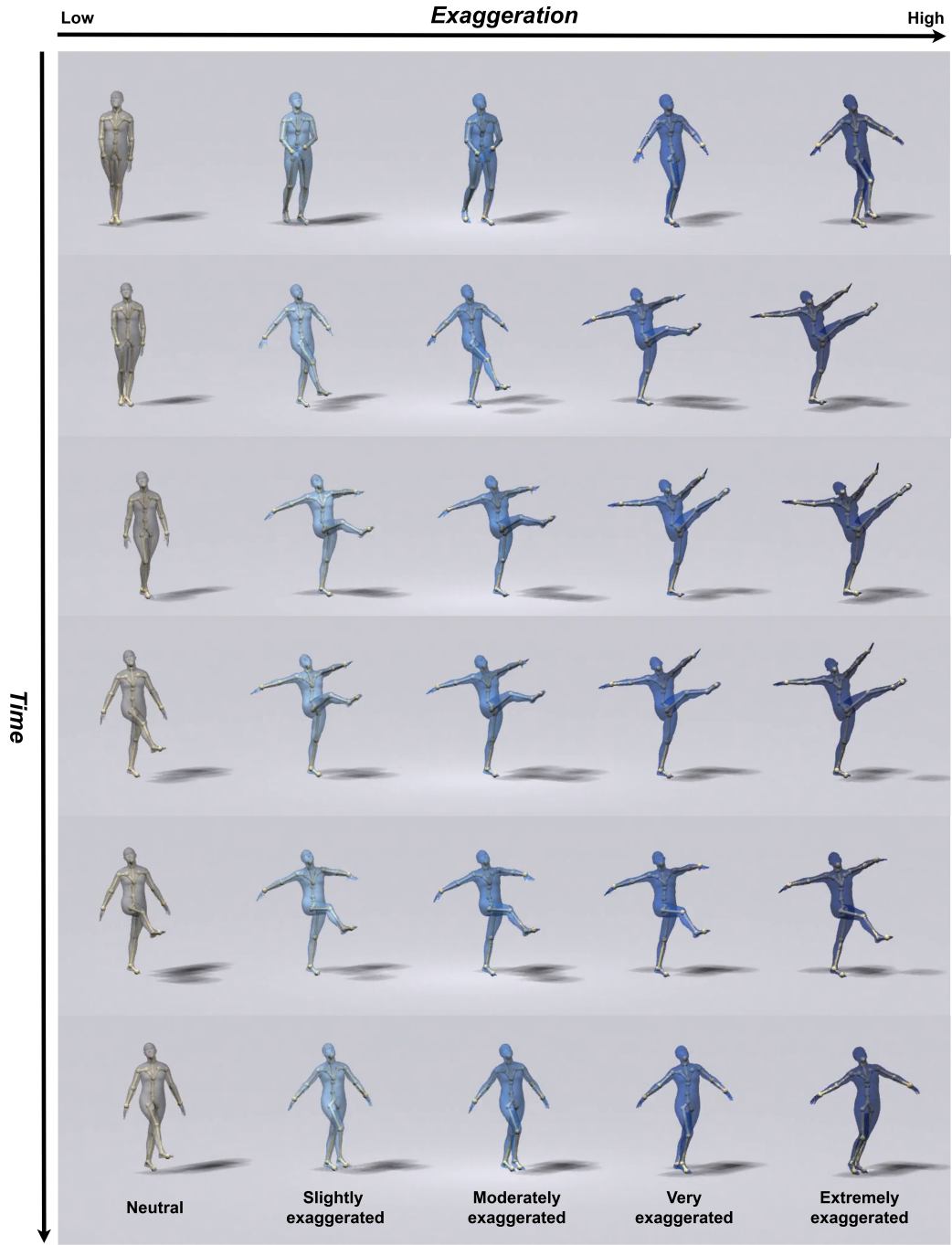}
        \vspace{-2pt}
        \caption{\textbf{Controllable exaggeration using our proposed supervised fine-tuning (SFT) approach.} The same kick motion is generated at progressively increasing exaggeration levels from left to right, with time progressing from top to bottom. Increasing exaggeration intensity through text prompt in the SFT approach produces progressively larger and more expressive poses while preserving the underlying neutral action.}
        \vspace{-13pt}
        \label{fig:result_controllable_exaggeration_sft}
    \end{figure}

    \begin{figure}[t]
        \vspace{-4pt}
            \centering
            \includegraphics[width=0.98\linewidth]{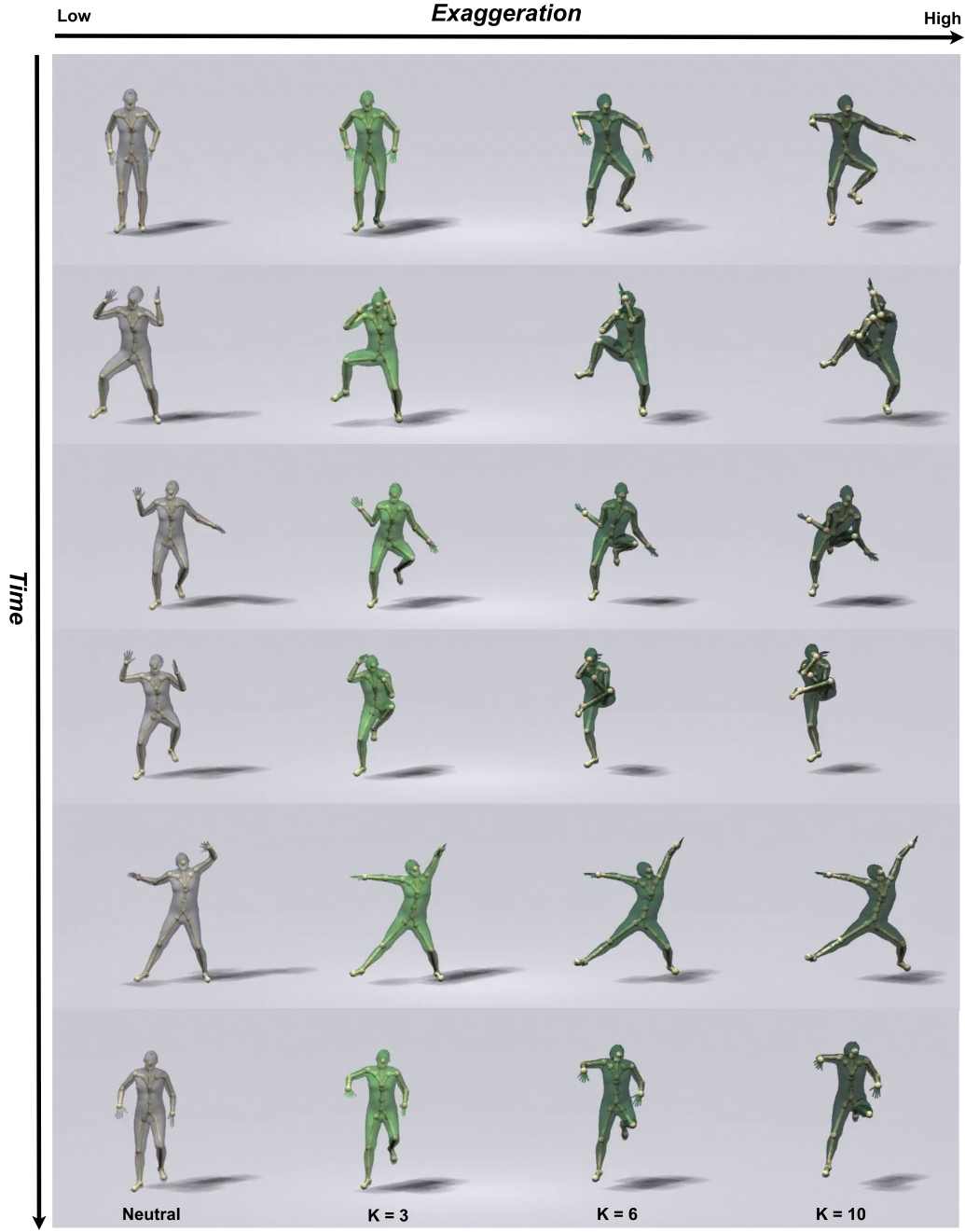}
            \vspace{-2pt}
           \caption{\textbf{Controllable exaggeration using our proposed training-free (TF) approach.} The same dance motion is generated at progressively increasing exaggeration levels from left to right, with time progressing from top to bottom. Increasing exaggeration intensity through the parameter $k$ in the TF approach produces progressively larger and more expressive poses while preserving the underlying neutral action.}
            \vspace{-13pt}
            \label{fig:result_controllable_exaggeration_tf}
    \end{figure}

    \begin{figure}[t]
        \vspace{-25pt}
            \centering
            \includegraphics[width=0.62\linewidth]{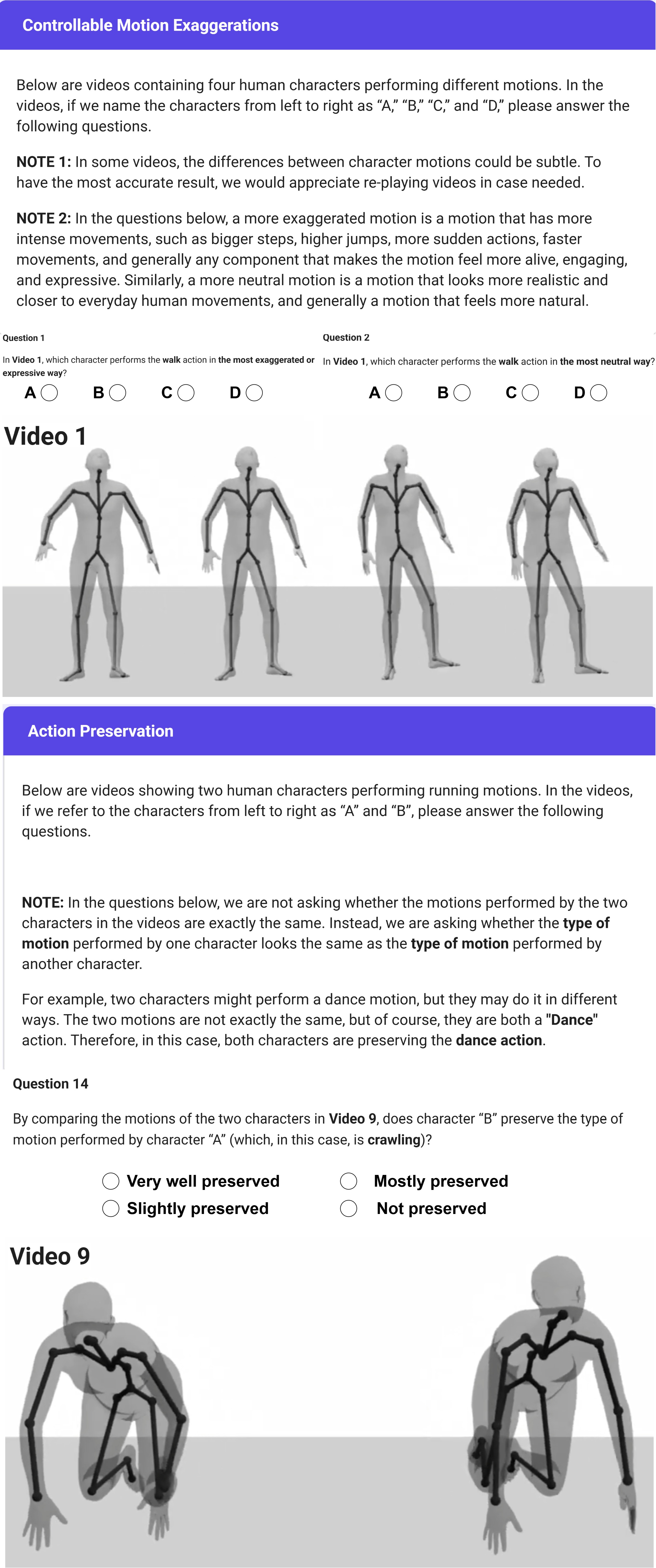}
            \vspace{-2pt}
            \caption{\textbf{Representative questions from our user study.} Top: for the controllable-exaggeration evaluation, participants viewed anonymized motions at different exaggeration levels and selected the most exaggerated and most neutral motions. Bottom: for the action-preservation evaluation, participants compared a neutral source motion with its exaggerated counterpart and rated how well the counterpart preserved the source action using a four-level response scale.}
            \vspace{-13pt}
            \label{fig:user_study_questionnaire}
    \end{figure}

\end{document}